\documentclass[preprint,12pt,authoryear]{elsarticle}

\usepackage{amssymb}
\usepackage[authoryear]{natbib}
\usepackage{subcaption}

\usepackage{tikz}
\usepackage{float}
\usepackage[edges]{forest}
\usepackage[T1]{fontenc}
\usepackage{lmodern}  
\usepackage[table]{xcolor}
\usepackage{makecell}
\usepackage{multirow}
\usepackage{booktabs}
\usepackage{amsmath}
\usepackage{url}
\usepackage{gensymb}
\usepackage{graphicx}
\usepackage{verbatim}
\usepackage{tabularx} 
\usepackage{algorithm}
\usepackage{algorithmicx}
\usepackage{algpseudocode}
\usepackage{minted}
\usepackage{listings}
\usepackage[utf8]{inputenc}
\RequirePackage{stfloats}
\usepackage{xurl}
\usepackage[colorlinks=true, linkcolor=blue, citecolor=blue, urlcolor=blue, hidelinks]{hyperref} 
\usepackage{adjustbox}

\def\tsc#1{\csdef{#1}{\textsc{\lowercase{#1}}\xspace}}
\tsc{WGM}
\tsc{QE}

\begin{document}

\begin{frontmatter}


\title{Integrated hybrid edge--cloud architecture for multi-robot and human-robot collaboration using ROS 2 – MQTT - FIWARE in Mediterranean Greenhouses}


\author[UAL-Inf]{Fernando Cañadas-Aránega}\corref{corr}\ead{fernando.ca@ual.es}
\cortext[corr]{Corresponding author}

\author[UAL-Inf]{Manuel Muñoz}\ead{mmr411@ual.es}

\author[UAL-Inf]{José C. Moreno}\ead{jcmoreno@ual.es}

\author[UAL-Eng]{José L. Blanco-Claraco}\ead{jlblanco@ual.es}


\affiliation[UAL-Inf]{
    organization={Department of Informatics, CIESOL, ceiA3, Universidad de Almería},
    addressline={Ctra. Sacramento s/n},
    city={Almería},
    postcode={04120},
    country={Spain}
}

\affiliation[UAL-Eng]{
    organization={Department of Engineering, CIESOL, ceiA3, Universidad de Almería},
    addressline={Ctra. Sacramento s/n},
    city={Almería},
    postcode={04120},
    country={Spain}
}


\begin{abstract}
The imperative to develop more sustainable agriculture demands a transition from isolated automation toward the deployment of multi-robot systems (MRS) in agrifood environments. However, Mediterranean greenhouse settings—characterized by narrow corridors, dense biomass, and structural metallic interference—pose significant challenges for robust and scalable communication between agents. Traditional robotic frameworks, such as ROS 2, frequently encounter node discovery issues and latency spikes due to dynamic obstacles, dense foliage, and other characteristic greenhouse elements, creating a critical bottleneck for real-time coordination. This paper proposes an innovative cloud-based hybrid architecture that establishes a two-way communication bridge between ROS 2, acting as an edge computing platform, and iVeg as a Decision Support System (DSS), using MQTT and the European FIWARE platform. The proposed framework enables seamless interoperability between fleets of multiple robots in environments with communication constraints, facilitating the synchronised exchange of high-level telemetry, point cloud data and farmer identification for collaboration, amongst other critical data. The architecture was validated in a high-fidelity simulation environment and subsequently tested in a real-world greenhouse scenario, demonstrating its ability to maintain persistent connectivity and data integrity under adverse network conditions. The results indicate that the integration of MQTT effectively eliminates information silos, providing a scalable and decentralised solution for managing complex robotic missions, which are executed locally via Edge Computing. This work sets a new methodological precedent for the concept of ‘Greenhouse Models as a Service’ (GMaaS), bridging the gap between low-level robotic control and high-level, cloud-based IoT decision-making.
\end{abstract}



\end{frontmatter}

\section{Introduction} 

Over the last few decades, greenhouse agriculture has consolidated as a pivotal strategy for enhancing food productivity while optimizing water and energy resource management. These advancements are driven, in part, by the progressive integration of automated systems and advanced control mechanisms in critical processes such as temperature, humidity, and ventilation management \citep{van2010optimal}. Within the field of automation, mobile robots designed for transportation, inspection, and operator assistance have emerged as essential agents in the transition toward digitized and connected agriculture \citep{hernandez2025reconfigurable}. Although greenhouses are structured environments, they pose significant challenges for autonomous mobility: narrow corridors that maximize cultivation area at the expense of manoeuvrability; terrains with non-linear behaviors—including slippage, heterogeneous compaction, or the presence of crop residues, or dynamic obstacles in general, and fluctuating environmental conditions \citep{canadas2024greenbot,canadas2026ros2}. These factors hinder navigation and necessitate robust control strategies capable of ensuring precise and safe motion even under high degrees of uncertainty.

Beyond the individual capabilities of a single platform, the true optimization of these agricultural environments lies in the deployment of Multi-Robot Systems. Shifting from single-agent operations to autonomous robot fleets allows for large-scale tasks—such as distributed harvesting or efficient monitoring—to be addressed cooperatively, thereby increasing system resilience and reducing execution times through operational parallelism \citep{das2025performance, MORENO2024100638}. However, coordinating multiple units within the confined spaces of Mediterranean greenhouses introduces a critical layer of complexity: the requirement for robust and constant data exchange. In these scenarios, communication management becomes a technological bottleneck; the high density of vegetation and structural interference demand network architectures capable of guaranteeing real-time agent synchronization \citep{11419082}.

In this context, the integration of the Internet of Things (IoT) emerges as a key paradigm for enabling interoperable and trustworthy robotic fleets, managing the communication complexity of distributed systems while fostering transparency through open data ecosystems \cite{khanh2022wireless}. Conceptually, IoT establishes a ubiquitous network of interconnected objects, including devices, sensors, actuators, and robotic agents, which converge into a digital ecosystem for bidirectional data exchange. This synergy not only bridges the physical greenhouse environment with virtual control layers but also enables integrated architectures capable of processing massive data streams across heterogeneous domains \cite{elijah2018overview}. While IoT deployment has transformed sectors such as smart cities \cite{zanella2014internet}, campus management \cite{chang2018campus}, and home automation \cite{handarini2020pembelajaran}, its application in autonomous vehicles and agricultural robotics \cite{munoz2020iot} represents currently the technological frontier for achieving field-deployable, trustworthy distributed autonomy. Standardizing robot orchestration via IoT protocols stands as a key strategy to deliver interpretable, contextual data from the field to human operators, transforming raw telemetry into active decision support.

This new approach requires the development of transversal protocols, interoperability, and close cooperation between services and systems. Currently, a significant portion of research focuses on cooperative robotic systems that enhance operational efficiency by combining mechanical precision with advanced perception and teamwork capabilities in shared environments \cite{lytridis2021overview}. However, this introduces additional design complexity: the need to capture, in real-time, the localization and status of each agent in both simulated and real-world environments. The optimal strategy for managing this information flow relies on wireless data exchange via the Internet, facilitating the integration of IoT architectures for real-time communication. This challenge is particularly critical within the technical atmosphere of a greenhouse \cite{munoz2020new}, where environmental complications must be emulated in high-fidelity simulators to validate communication through shared data clouds prior to physical implementation. In this sense, intelligent robotic agriculture transcends mere automation by integrating Big Data services, Cloud Computing, and Farm Management Information Systems (FMIs) powered by Artificial Intelligence \cite{munoz2020iot}. Nevertheless, integrating these services in agricultural environments remains a complex task requiring high-performance connectivity to ensure precision in robot navigation. To overcome these barriers, the use of cloud platforms providing open data management systems that easily integrate heterogeneous devices enables the evolution toward "Greenhouse Models as a Service" (GMaaS) models, optimizing system scalability and flexibility.

To address the critical connectivity and data management challenges previously outlined, this paper contributes to the digitalization of the agricultural sector by presenting a cloud-based architecture specifically engineered for robotic fleets operating in dynamic environments. The fundamental novelty of this proposal lies in providing a robust communication framework for high-complexity scenarios, such as Mediterranean greenhouses. In these settings, dense biomass and extremely narrow aisle configurations cause severe degradation of conventional wireless signals and constant sensory occlusions, which render current multi-robot communication solutions practically unfeasible. The proposed approach overcomes these limitations by transforming visual detections of operators and other robots into lightweight geometric metadata. This information is synchronized via a bidirectional ROS 2-MQTT bridge with the iVeg platform \cite{macenski2022ros2,munoz2020iot}. Fundamentally, this architecture enables a cooperative safety mechanism where robots maintain shared spatial and collaborative awareness. By broadcasting the real-time coordinates of detected farmers, peer agents can execute pre-emptive deceleration reducing their velocity by 50\% even when the obstacle is situated outside their immediate field of view. By replacing bandwidth-intensive data streams with optimized JSON messages, the architecture ensures the persistence of fleet-wide situational awareness, even in areas with minimal connectivity \cite{vsvec2022multi}. As detailed in the literature review, no previous work natively integrates ROS 2 with an MQTT cloud infrastructure to guarantee distributed safety and autonomy in highly occluded, collaborative protected cultivation environments.

Based on the technical literature review, the gaps this work intends to bridge have been identified. Consequently, the primary contributions of this article are summarized as follows:

\begin{itemize}
    \item \textit{Multi-Robot Architecture for Greenhouses:} This work presents the development of a high-fidelity simulation environment and real-world trials that enable the integration of cooperative robotic fleets. This architecture ensures that interoperability among heterogeneous agents translates into a shared world model, allowing robots to perceive human presence through inter-agent data exchange.

    \item \textit{Human-Robot Interaction:} The proposed architecture guarantees that interoperability among heterogeneous agents results in a shared world model including farmers, enabling robots to proactively perceive human presence through the exchange of localised agent data.
    
    \item \textit{Hybrid Network Architecture for Critical Connectivity:} An innovative cloud-based solution utilizing MQTT and FIWARE is proposed to optimize real-time data handling. This demonstrates that lightweight messaging is the optimal solution for maintaining persistent synchronization in greenhouse infrastructures where network stability is frequently compromised by biomass.
    
    \item \textit{Bidirectional Safety Bridge between FIWARE - MQTT - ROS 2:} This work establishes a robust bidirectional communication bridge between ROS 2 and MQTT in accordance with the FIWARE standard. For the first time, this integration is utilized to implement closed-loop safety commands from the cloud, enabling automatic speed modulation based on collective intelligence.
\end{itemize}

The rest of the article is organised as follows. Section \ref{sec: RW} presents a review of the state of the art in current multi-robot systems in agriculture and the IoT technology used. The materials and methods used are described in Section \ref{sec: MM}. The detailed description of the architecture is explained in Section \ref{sec: Robot}. Section \ref{sec: result} discusses the description of the experiments carried out. Finally, Section \ref{sec: conclusion} is devoted to some conclusions.

\section{Related Work} \label{sec: RW}

To provide a comprehensive context for the proposed architecture, this section reviews the state of the art in collaborative and cooperative agricultural robotics, IoT-robotic integration, and interoperable data models. By analyzing these domains, the technical gaps, specifically regarding communication persistence and cross-platform synchronisation, are highlighted to show what this work seeks to overcome.

\subsection{Multi-Robot cooperation in Agriculture}

Cooperative robotics has emerged as a disruptive paradigm for addressing the labor-intensive nature of modern farming, particularly within protected cultivation environments. \cite{farella2024agricultural} underscored that while the potential for efficiency is high, the deployment of Multi-Robot Systems in confined spaces faces significant constraints regarding spatial navigation and real-time synchronization. In this context, heterogeneous fleets comprising robots with varying kinematic constraints have gained significant research attention. For instance, \cite{fawakherji2019crop} demonstrated that collaborative sensing enriches data management for crop classification. Building on this foundation, \cite{roldan2016heterogeneous} proposed a decentralized coordination framework for multi-robot harvesting, emphasizing the need for an interoperable communication layer to bridge the gap between low-level control via ROS 2 and high-level decision-making (Cloud-IoT). Furthermore, the work of \cite{lytridis2021overview} suggests that the future of greenhouse robotics lies in "Swarm-to-Cloud" architectures, where multiple agents utilize lightweight protocols such as MQTT to maintain a shared world model despite signal attenuation caused by metallic greenhouse structures. Despite these advances, a clear research gap remains regarding the provision of a unified, real-time communication bridge that ensures the scalability of these cooperative tasks within Mediterranean-style infrastructures.

This challenge is further exacerbated by the hostile nature of the greenhouse environment for wireless communications. Research on signal propagation in dense crops, such as the work by \cite{ferentinos2017wireless}, demonstrates that high foliage density and humidity levels act as physical barriers, causing severe multi-path fading and energy absorption that degrade link quality. Beyond ground-based platforms, \cite{boursianis2022internet} explored the integration of Unmanned Aerial Systems (UAS) for mission-critical monitoring, employing heuristic techniques such as particle swarm optimization to manage complex trajectories in greenhouses. This physical instability directly impacts the transport layer in the Open Systems Interconnection (OSI) model \citep{antonelli2013interconnected}; studies on IoT protocols in precision agriculture by \cite{hernandez2025reconfigurable} point out that although protocols like MQTT are lightweight, their reliance on TCP leads to critical delays and constant retransmissions in environments with high packet loss due to plant obstacles, thereby compromising real-time data delivery. Recent studies by \cite{miele2025distributed} highlight that coordination within greenhouses is severely hampered by the absence of reliable communication middleware capable of handling the high data throughput required by vision-based learning approaches. Therefore, the transition from individual automation to collective intelligence remains a technological bottleneck. 

\subsection{Deep Learning and Machine Learning Vision Model in Agriculture}

In the domain of autonomous agricultural vehicles, deep learning-based perception has transitioned from generic monitoring to mission-critical multi-agent safety. Advanced object detection models, particularly the YOLO (You Only Look Once) family, have become the standard for enhancing detection accuracy in unstructured environments \citep{wang2024yolov10}. However, in Mediterranean greenhouses, the extreme density of biomass and narrow aisle configurations pose a dual challenge: they create constant sensory occlusions for local vision systems and simultaneously degrade wireless signals, hindering the exchange of raw visual data. Recent studies emphasize that while YOLOv8 and YOLOv10 provide superior identification of personnel and heterogeneous robot morphologies \citep{canadas2024greenbot}, their utility in multi-robot coordination is often bottlenecked by the inability of standard networks to transmit high-bandwidth information in such occluded settings.

To address these challenges, recent research has shifted toward perception-driven communication architectures. Instead of transmitting heavy video streams, optimized frameworks focus on the extraction of lightweight geometric metadata specifically 2D bounding box coordinates and class probabilities to maintain real-time fleet synchronization \citep{luo2016distributed}. While heavy architectures like ResNet or VGG16 suffer from high computational latency \citep{chen2019deep}, the integration of attention mechanisms, such as the Convolutional Block Attention Module (CBAM), has significantly improved bounding box regression accuracy, allowing for precise distance estimation between cooperative agents and farmers even under non-linear lighting \citep{he2019bounding}. The state-of-the-art now suggests that bridging these ROS 2-based local detections with MQTT-based cloud infrastructures is the only viable path for distributed autonomy in dense crops \citep{chan2025real}. This approach enables a shared world-model where local perception enriches global awareness without saturating the constrained bandwidth of protected cultivation environments.

\subsection{IoT Integration in Multi-Robot Systems in Greenhouses}

The convergence of robotics and IoT has fostered the concept of the Internet of Robotic Things (IoRT), where local processing capabilities are augmented by cloud-based resources. Numerous studies have explored the use of IoT to enhance the control and monitoring of autonomous agents, initially employing short-range technologies such as Bluetooth \citep{khoje2016robotic}, Wi-Fi \citep{madkar2016robot}, ZigBee \citep{ZigBee}, and Wireless Sensor Networks (WSN) \citep{mishi2017multiple}. However, these solutions often exhibit limitations in expansive agricultural settings due to restricted coverage ranges and high sensitivity to physical obstructions. To overcome these barriers, recent literature has shifted toward multilayered systems that integrate distributed Edge-Computing nodes with scalable IoT Cloud platforms. These advanced architectures leverage decentralized middleware and low-latency communication protocols to orchestrate autonomous robots, enabling high-throughput data orchestration in complex domains such as precision agriculture \citep{ramesh2020design}, perimeter surveillance \citep{telkar2020iot}, and emergency rescue operations \citep{narendra}.

Despite these developments, the current state of the art reveals that interoperability between heterogeneous robots and sensors remains an open challenge, particularly in scenarios demanding real-time data management. Recent research by \cite{ojha2021internet} suggests that conventional ROS 2 protocols (based on DDS) frequently encounter node discovery issues in Wide Area Networks (WAN), which has motivated the adoption of lightweight messaging protocols such as MQTT. However, the implementation of an efficient bidirectional bridge between the ROS 2 ecosystem and cloud-distributed MQTT brokers specifically tailored for environments with high humidity and structural occlusion, such as greenhouses remains a critical area of exploration.

\subsection{Data Models and Standards in IoT}

Data interoperability is a key concern in IoT. Works such as \cite{INTER-METH} and \cite{fortino2020internet} emphasize the importance of standardized data models to facilitate the integration of heterogeneous platforms. However, existing proposals often lack validation in real-world agricultural environments or fail to address real-time data integration between robots and sensors. Models like OMA NGSI have gained attention for their ability to ensure semantic and syntactic interoperability, but their specific application in collaborative robotics remains limited \citep{kovacs2016standards}.

Studies such as that of \cite{munoz2020new} highlight the importance of developing IoT solutions for integrating data generated by sensors in greenhouses, demonstrating the practical application of interoperability frameworks in controlled environments. By the effective processing of sensor data and the use of robotics, these solutions address farmers' needs by combining the precision and power of robotic platforms with the flexibility of cloud-based communication architectures, even in complex scenarios such as greenhouses \citep{corista2018iot}. However, existing proposals often lack validation in agricultural environments or do not address real-time data integration between robots and sensors.

\section{Materials and methods} \label{sec: MM}

\begin{figure}[!ht]
    \centering
    \includegraphics[width=0.4\linewidth]{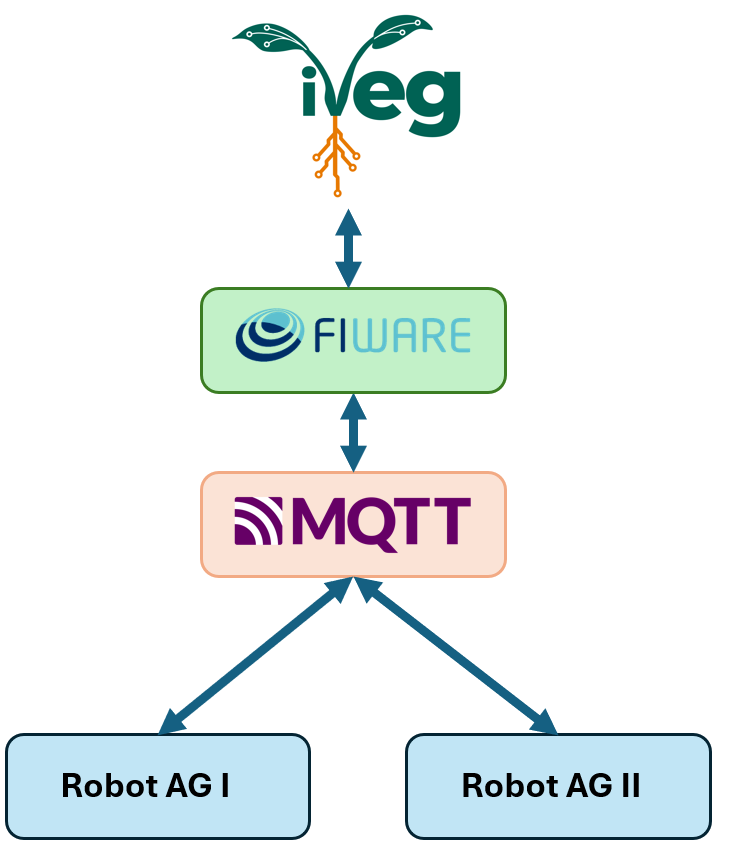}
    \caption{General system block diagram}
    \label{fig:1}
\end{figure}

In this study, a robust multi-robot communication architecture has been developed to integrate robotic agents operating under ROS 2 Humble Hawksbill through the MQTT protocol and the FIWARE ecosystem \citep{macenski2022ros2, mqtt, FIWARE}. This architecture is specifically designed to facilitate bidirectional data exchange within Mediterranean greenhouses—environments characterized by high signal attenuation and latency spikes caused by dense biomass and dynamic obstacles. The high-level system topology is illustrated in Figure \ref{fig:1}. The communication pipeline is structured as follows: robotic nodes publish telemetry and environmental data via MQTT to the FIWARE Orion Context Broker, which serves as the backbone for the iVeg platform. This integration establishes a scalable cloud-based application layer where high-level decision-making algorithms aggregate multi-source data to orchestrate coordinated robotic actions. The following subsections detail the materials and methodologies employed in the design and validation of this architecture

\subsection{Materials}

This section describes the materials used, beginning with a description of the greenhouse and the real robots, and 3D models used for the simulations.

\subsubsection{AgroConnect greenhouse facilities}

\begin{figure}[!ht]
    \centering
    \includegraphics[width=8cm]{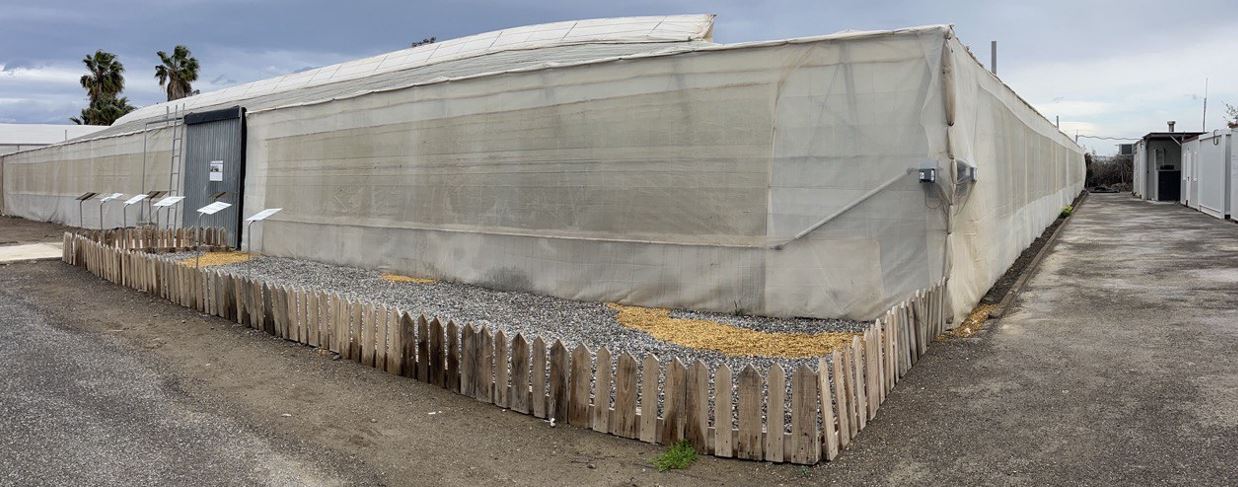} \centering
    \caption{Greenhouse outdoor}
    \label{fig:sub1}
\end{figure}

The experimental trials were conducted at the AgroConnect facilities (funded by the Ministry of Science, Innovation, and Universities and the European Regional Development Fund - FEDER, grant 2019) located in Almería, Spain ($36^{\circ}50' \text{ N}, 2^{\circ}24' \text{ W}$) (Figura \ref{fig:sub1}). 

\begin{figure}[!ht]
   \includegraphics[width=9cm]{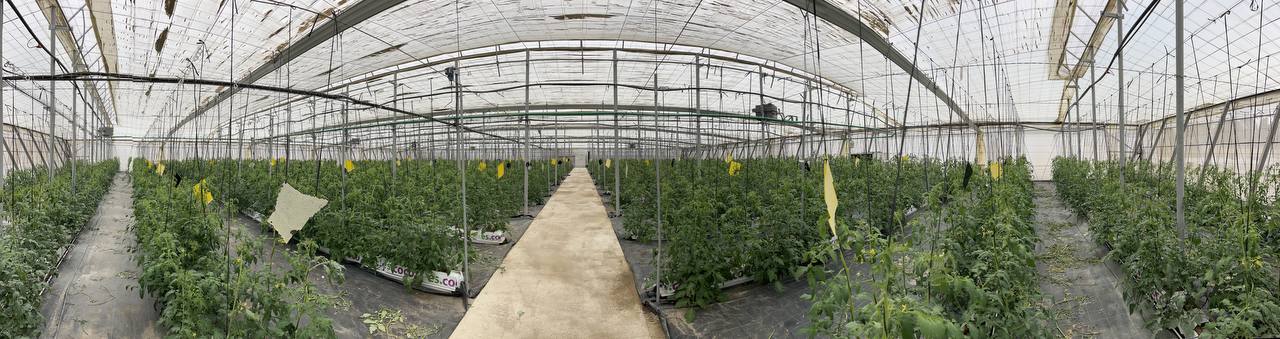} \centering
   \caption{Greenhouse indoor}
   \label{fig:sub2}
\end{figure}

The site is situated at an elevation of 3 m above sea level, featuring a consistent 1\% northern slope across the terrain. The testing grounds consist of a 1,850 m² Mediterranean-style greenhouse (locally known as ''raspa y amagado``), characterised by a robust steel structural frame and a high-density polyethene cladding. From a robotics perspective, the facility is designed around a 2 m wide central corridor that acts as the primary navigation artery, connecting to eleven lateral aisles on each side (Figure \ref{fig:sub2}). From the central corridor, 22 corridors branch off to the north and south; those to the north are 12.5 m long, whilst those to the south are 22.5 m long, all of them 2 m wide without taking plants into account.

The facility hosts a commercial tomato crop (\textit{Lycopersicon esculentum}) cultivated in coconut fiber (coir) bags arranged in strict north-to-south rows. Advanced climate control infrastructure including zenithal and lateral natural ventilation, an integrated HVAC network, CO$_2$ enrichment, and high-pressure humidification systems maintains optimal agronomic conditions.

\subsubsection{Simulator for AgroConnect greenhouse}

\begin{figure}[!ht]
\centering
\includegraphics[width=7cm]{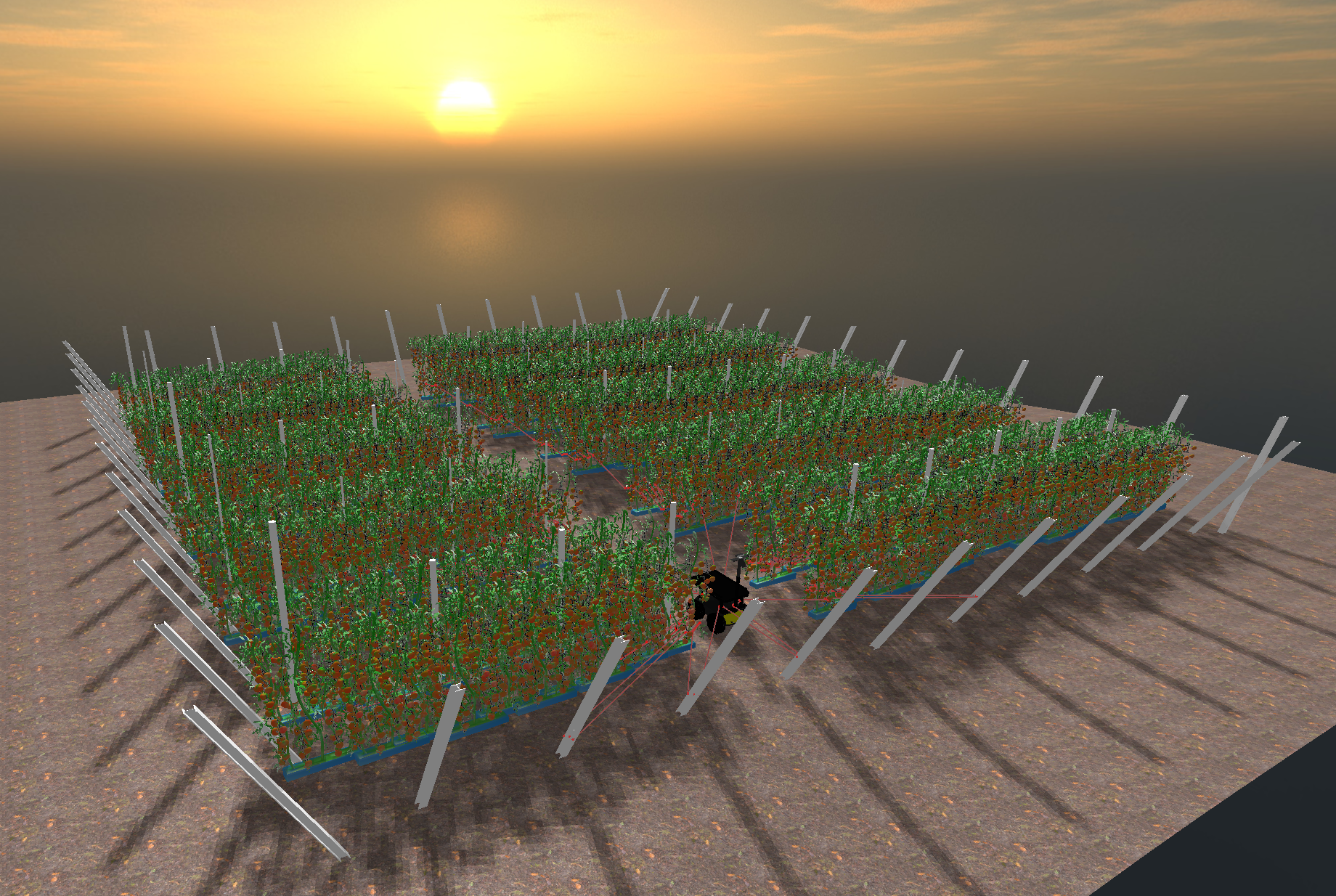} \centering
\caption{Complete 3D greenhouse model \citep{Canadasifac2026}}
\label{fig:Model3D1}
\end{figure}

To validate the safety and interoperability of the multi-robot coordination algorithms before physical deployment, the proposed framework was first evaluated within a high-fidelity simulation environment. It utilised a detailed 3D digital twin of the AgroConnect greenhouse, originally developed in SolidWorks and validated for autonomous navigation in \citep{aranega2024navegacion} (Figure \ref{fig:Model3D1}).

\begin{figure}[!ht]
    \centering
    \includegraphics[width=0.25\textwidth]{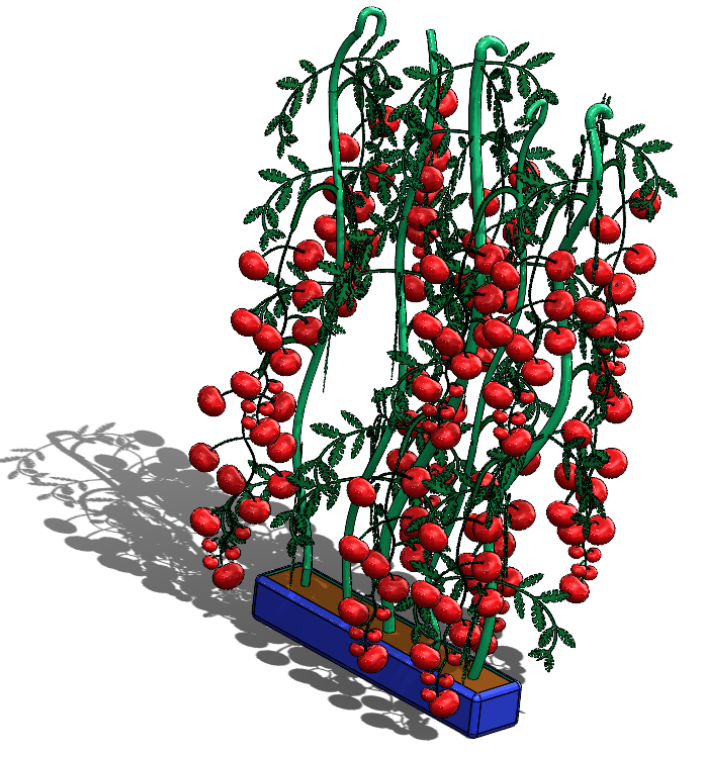}
    \caption{3D Tomato plant model}
    \label{fig:5}
\end{figure}

This model meticulously replicates the architectural complexity of the real facility, including primary structural columns and diagonal reinforcement elements, faithfully represented according to their real-world spatial configuration. Furthermore, a high-resolution 3D botanical model of the pear-tomato plant (\textit{Lycopersicon esculentum}) was integrated (see Figure \ref{fig:5}).

Both the vegetation and the structural pillars were positioned within the virtual world to strictly match their actual coordinates in the experimental site. For the purpose of this study, a $20 \times 20$ m virtual testbed was constructed, featuring five corridors separated by 4 m. Each lane is flanked by the aforementioned hydroponic tomato crop, effectively defining the navigation corridors for the multi-robot fleet. This environment accurately replicates the characteristic geometry and occlusion patterns of Mediterranean greenhouses, providing a proper base for validation of autonomous navigation and coordination algorithms previously established in the literature \citep{canadas2026ros2}.

\subsubsection{Fleet of Intelligent Robotic Agents}

The experimental multi-robot fleet comprises two distinct mobile platforms engineered to act as transparent, internet-connected agents. These units, the mobile robots AgriCobIoT I \citep{moreno2022modelado} and AgriCobIoT II \citep{canadas2024greenbot} are integrated into a cloud-based architecture through localized fog and edge computing layers. This design enables the robots not only to execute local manoeuvres autonomously but also to expose their internal edge-perception states to human supervisors in real time.

\begin{figure}[!ht]
    \centering
    \subfloat[3D Model Robot AGII \label{fig:AGIIm}]{
        \includegraphics[width=0.5\textwidth]{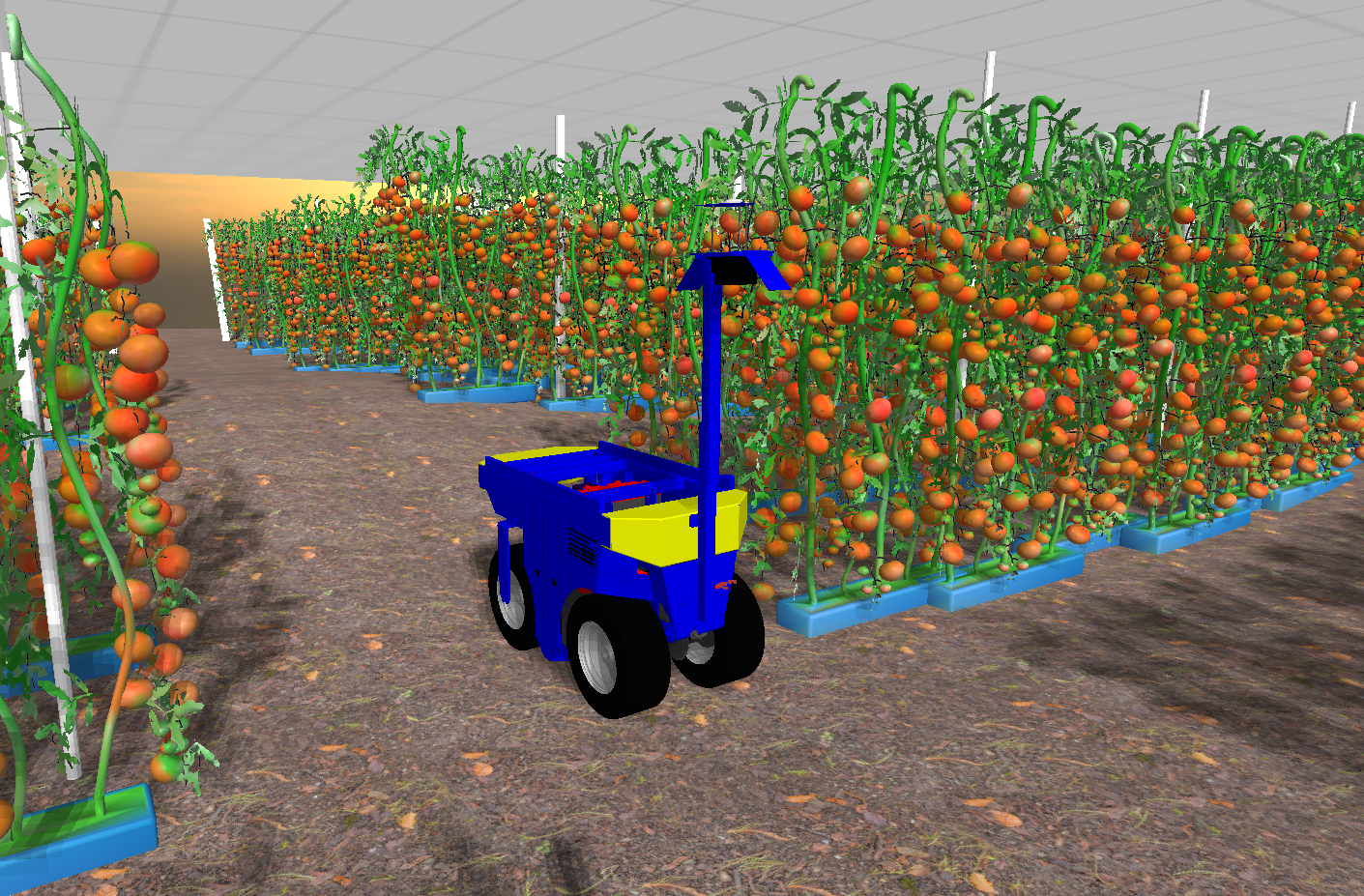}
    }
    \subfloat[Real Robot AGII \label{fig:AGIIr}]{
        \includegraphics[width=0.41\textwidth]{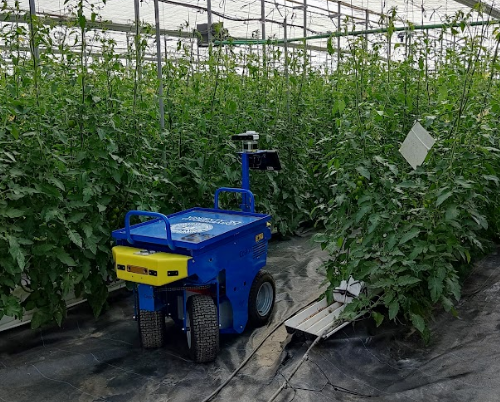}
    }
    \caption{Real 3D models and AGII used}
    \label{fig:RobotAGII}
\end{figure}

\begin{itemize}
    \item AgriCobIoT I (AGI)

Based on the commercial Husky platform, AGI utilizes a differential drive system capable of executing zero-radius turns, an essential kinematic requirement for navigating the abrupt transitions between the central corridor and the narrow cultivation aisles. To support human workers, the platform features a custom-built ergonomic superstructure tailored for automated crate transportation. This architecture includes four central support pillars and a removable, rail-guided mechanism engineered for standard agricultural boxes. Safety and proximity sensing are managed by base-level ultrasonic arrays, while a specialized rear mast houses a height-adjustable perception payload dedicated to human detection and tracking during collaborative tasks \citep{moreno2022modelado,Greensys}. The physical system is depicted in Figure \ref{fig:RobotAGI}b. Its virtual counterpart leverages a computationally optimized \texttt{.dae} mesh balanced with simplified collision primitives to ensure stable rigid-body physics simulation (Figure \ref{fig:RobotAGI}a).

\begin{figure}[!ht]
    \centering
    \subfloat[3D Model Robot AGI \label{fig:AGIIm2}]{
        \includegraphics[width=0.45\textwidth]{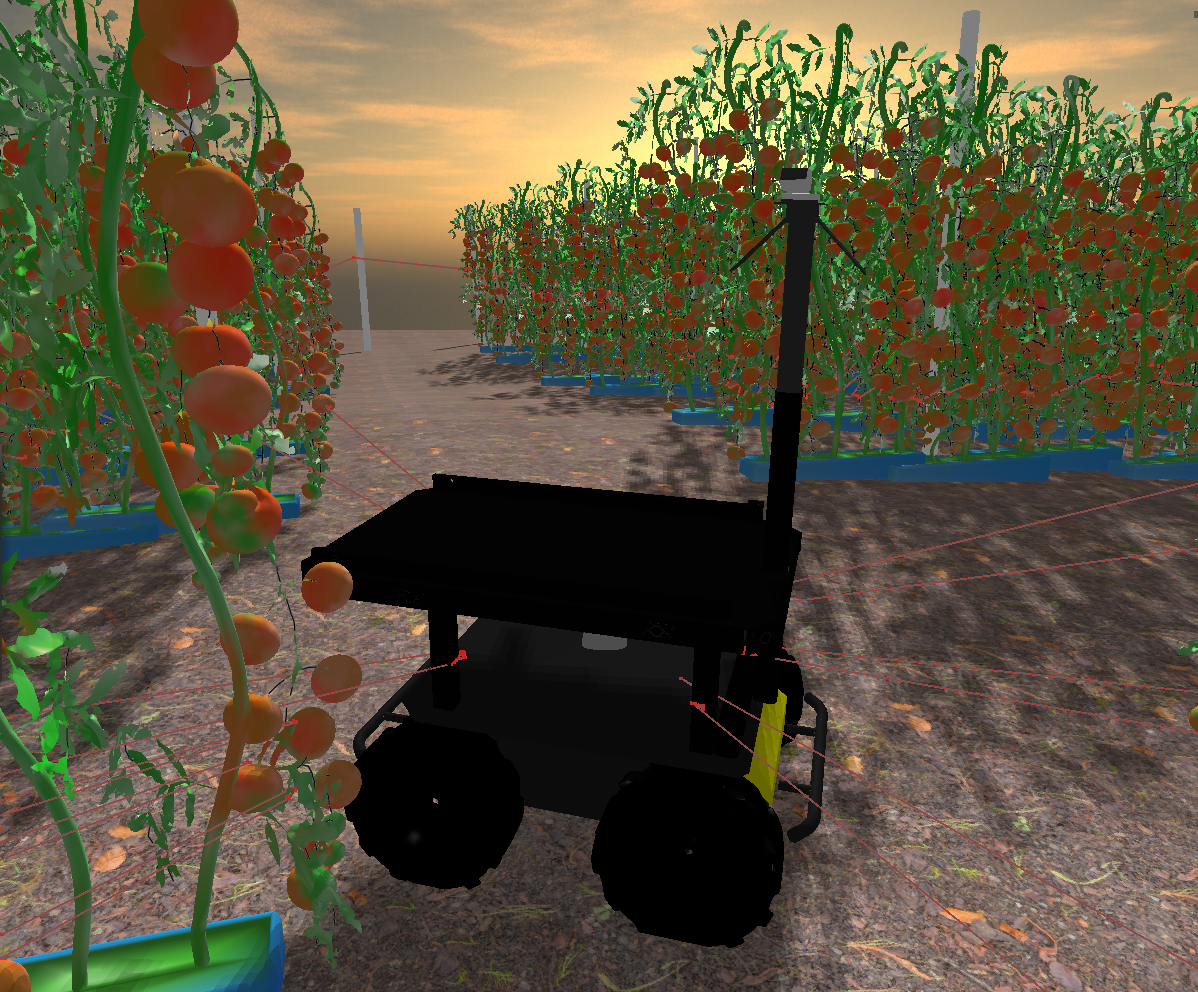}
    }
    \subfloat[Real Robot AGI \label{fig:AGIIr2}]{
        \includegraphics[width=0.36\textwidth]{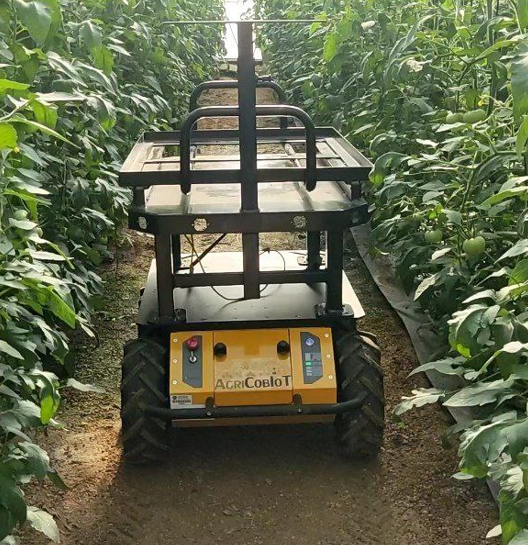}
    }
    \caption{Real 3D models and AGII used}
    \label{fig:RobotAGI}
\end{figure}

For the purpose of this study, the platform is equipped with the following core hardware layer:

\begin{itemize}
    \item \textit{HITTSON Industrial Control Unit:} A high-performance onboard computer that serves as the agent's edge-computing core. It features an integrated Wi-Fi 6 (802.11ax) module, enabling low-latency, bidirectional telemetry exchange with the local greenhouse router.
    \item \textit{Intel RealSense D435 RGB-D Camera:} An active stereo depth sensor utilized as the primary vision engine. Crucially, this sensor does not stream heavy, privacy-invasive raw video feeds to the cloud. Instead, it feeds high-resolution spatial data directly into the local edge deep-learning model, extracting precise 3D bounding boxes ($XYZ$ coordinates) and model confidence levels to generate lightweight, human-interpretable geometric metadata.
    \item \textit{Velodyne VLP16}: A 16-channel Time-of-Flight (ToF) laser scanner featuring a $360^\circ$ horizontal field of view. Characterized by its compact form factor and high sampling rate (up to $300,000$ points/s), this sensor is pivotal for Simultaneous Localization and Mapping (SLAM) and real-time obstacle and ground detection \citep{canadas2026greenseg}. Within this study, it provides the high-fidelity spatial data required to validate navigation safety and maneuverability within the constrained aisles of a greenhouse environment.
    \item \textit{Taobotics 9-axis IMU:} An Inertial Measurement Unit fused with wheel odometry. By outputting real-time linear accelerations and angular velocities, it guarantees robust pose estimation across the irregular, sloped, and low-traction soils of Mediterranean greenhouses, ensuring that the spatial explanations sent to the user interface remain physically accurate.
\end{itemize}

These sensors are also simulated in the virtual environment.\\

    \item AgriCobIoT II (AGII)

Designed at the University of Almería, AGII provides a complementary kinematic profile to the fleet \citep{canadas2024greenbot}. It utilizes an Ackermann steering geometry with a nominal 2~m turning radius. While structurally constrained by this steering mechanism, AGII compensates with a heavy-duty payload capacity of 150~kg, making it optimal for large-scale crop hauling. Mirroring AGI's layout, its chassis optimizes sensor placement and crate allocation to streamline harvesting workflows. A secondary rear mast houses an identical perception payload to maintain fleet-wide data symmetry during autonomous tracking operations. The physical realization and corresponding 3D mesh of AGII are shown in Figure \ref{fig:RobotAGII}.
    
The same sensors used in the physical model have been also incorporated into the 3D model as well.\\

\end{itemize}
\subsubsection{MultiVehicle Simulator (MVSim)}

\begin{figure}[!ht]
   \includegraphics[width=11cm]{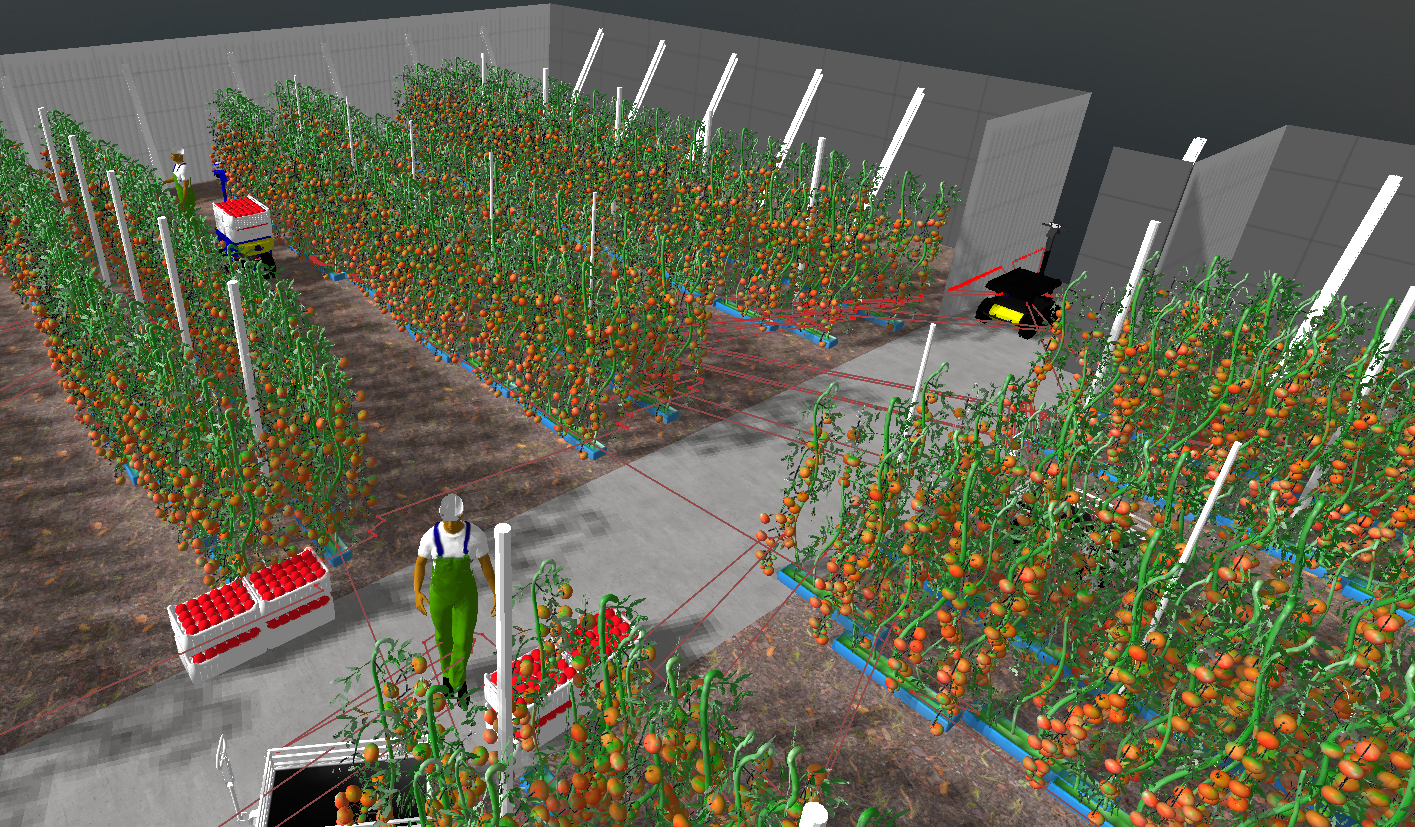} \centering
   \caption{Greenhouse environment with robots at MVSim}
   \label{fig:mvsim}
\end{figure}

For the pre-operational assessment, a simulation is created in the MultiVehicle Simulator (MVSim) \citep{blanco2023multivehicle}, an environment that supports real-time simulation of multiple vehicles or robots in simple and complex settings, and it runs in ROS. MVSim uses realistic physics-based friction models to accurately simulate the interaction between the tires and the ground. The simulator supports using the most common modern mobile robotics and autonomous vehicle research sensors, such as RGB-D cameras or 2D and 3D LiDAR scanners. All depth-related sensors can accurately measure distances to user-supplied 3D models to define elements of a custom environment using acceleration based on the Graphics Processing Unit (GPU). When working with ground vehicles, the simulator is very efficient, as a simplified 2D physics engine is used for body-to-body collisions and to resolve wheel-ground interaction forces separately, used in previous works such as \citep{canadas2026ros2}. By importing the exact \texttt{.dae} structures of AGI and AGII, MVSim acts as an effective auditing tool for the explainable safety framework. As illustrated in Figure \ref{fig:mvsim}, the simulation world was populated with lifelike 3D human models (representing greenhouse personnel), standard tomato crates, and manual harvesting trolleys.

\subsection{Methods}

This section describes the methods and algorithms used in the development of multi-robot system, which details the ROS 2 middleware used, the farmer recognition methodology, the communication protocols and the multisensor network that has been built.

\subsubsection{Robot Operating Systems 2 (ROS 2)}

The core of the robotic software architecture is built upon ROS 2 Humble Hawksbill, which serves as the distributed middleware for high-level task orchestration and hardware abstraction. Unlike its predecessor, ROS 2 leverages the Data Distribution Service (DDS) standard to provide the Quality of Service (QoS) policies essential for maintaining reliable communication in the unpredictable conditions of a greenhouse \citep{macenski2022ros2}. In this multi-robot framework, ROS 2 facilitates inter-agent interoperability by utilizing standardized message interfaces that allow heterogeneous platforms to share a unified world model. To ensure scalability and prevent data interference, each robot operates within a dedicated namespace, allowing the MQTT bridge to aggregate local perception data into a global fleet management layer without cross-talk. Furthermore, the system employs the \texttt{tf2} transform library to synchronize the spatial coordinate frames of all agents, transforming local YOLOx detections into a shared global map. This integration of decentralized communication and real-time spatial synchronization transforms individual autonomous units into a cohesive, perception-aware fleet capable of the complex cooperative navigation and collaborative human-robot interaction required in dense agricultural environments \citep{macenski2020marathon}.

\subsubsection{Farmers and Robots Recognition with YOLOx}

\begin{figure}[!ht]
    \centering
    \includegraphics[width=0.75\textwidth]{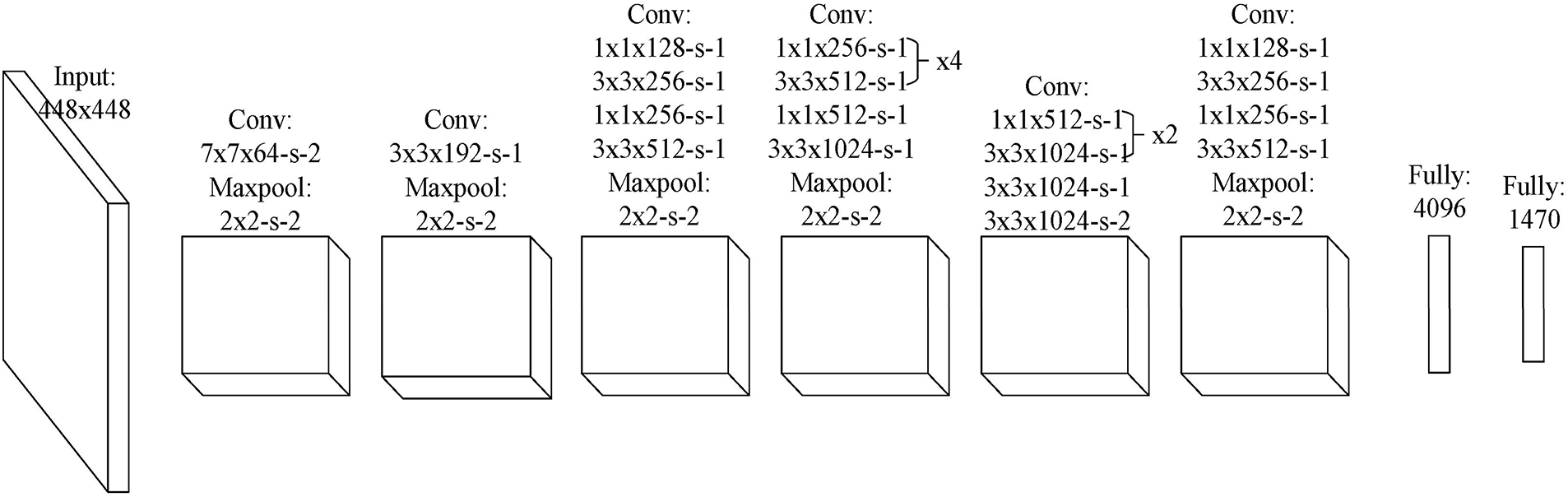}
    \caption{The basic YOLO network structure \citep{yayla2025embedded}}
    \label{fig:YOLO}
\end{figure}

To facilitate transparent agent identification within the occluded greenhouse environment, a vision-based object detection system utilizing the YOLOx framework\footnote{YOLOx-ROS repository: \url{https://github.com/Ar-Ray-code/YOLOX-ROS/tree/humble}} (an adaptation of YOLOv10 for ROS 2) was implemented and optimized for the Humble distribution. Unlike multi-stage detection methods, YOLOx processes images in a single neural network pass with AI. This edge-computing approach ensures the ultra-low latency required for proactive safety maneuvers and dynamic obstacle avoidance. From an explainability standpoint, the core strength of YOLO lies in its holistic image processing; it evaluates contextual information to drastically reduce false-positive detections, laying a reliable foundation for human-robot trust. The architectural structure of the deployed YOLO network is illustrated in Figure \ref{fig:YOLO} \citep{yayla2025embedded}.

This versatility, combined with its refined architecture and a user-friendly API, makes YOLO a powerful choice for field-deployable computer vision tasks \citep{xu2024survey}. YOLO reframes object detection as a single regression problem, directly predicting spatial bounding boxes and class probabilities. The core of its mathematical formulation is the loss function, which is a weighted sum of three main components: localization loss, confidence loss, and classification loss \citep{zhang2025method}. This unified approach allows the model to predict bounding boxes and class probabilities simultaneously in real-time. The total loss function is defined in Eq. (\ref{eq: YOLO}).

\begin{equation}
\begin{aligned}
\text{Loss} &= \lambda_{\text{coord}} \sum_{i=0}^{S^2} \sum_{j=0}^{B} \mathbb{1}_{ij}^{\text{obj}} \left[ (x_i - \hat{x}_i)^2 + (y_i - \hat{y}_i)^2 \right] \\
&+ \lambda_{\text{coord}} \sum_{i=0}^{S^2} \sum_{j=0}^{B} \mathbb{1}_{ij}^{\text{obj}} \left[ (\sqrt{w_i} - \sqrt{\hat{w}_i})^2 + (\sqrt{h_i} - \sqrt{\hat{h}_i})^2 \right] \\
&+ \sum_{i=0}^{S^2} \sum_{j=0}^{B} \mathbb{1}_{ij}^{\text{obj}} (C_i - \hat{C}_i)^2 \\
&+ \lambda_{\text{noobj}} \sum_{i=0}^{S^2} \sum_{j=0}^{B} \mathbb{1}_{ij}^{\text{noobj}} (C_i - \hat{C}_i)^2 \\
&+ \sum_{i=0}^{S^2} \mathbb{1}_{i}^{\text{obj}} \sum_{c \in \text{classes}} (p_i(c) - \hat{p}_i(c))^2
\end{aligned}
\label{eq: YOLO}
\end{equation}

The parameters and components in this equation are interpreted as follows:

\begin{itemize}
    \item $S^2$: The number of grid cells in the input image (e.g., a $7 \times 7$ grid means $S^2 = 49$).
    \item $B$: The number of bounding boxes that each grid cell is configured to predict.
    \item $\mathbb{1}_{ij}^{\text{obj}}$: An indicator function that is 1 if the $j$-th bounding box in the $i$-th grid cell is responsible for detecting an object, and 0 otherwise.
    \item $\mathbb{1}_{ij}^{\text{noobj}}$: An indicator function that is 1 if the $j$-th bounding box in the $i$-th grid cell does not contain any object.
    \item $x_i, y_i, w_i, h_i$: The predicted bounding box coordinates (originating at the lower left vertex) and dimensions.
    \item $\hat{x}_i, \hat{y}_i, \hat{w}_i, \hat{h}_i$: The ground truth (actual) bounding box coordinates (originating at the lower left vertex) and dimensions.
    \item $C_i$: The predicted confidence score for a bounding box.
    \item $\hat{C}_i$: The ground truth confidence score. This is 1 if the box contains an object and 0 otherwise.
    \item $p_i(c)$: The predicted class probability for class $c$.
    \item $\hat{p}_i(c)$: The ground truth class probability.
    \item $\lambda_{\text{coord}}$ (Lambda Coordinate): This crucial hyperparameter weighs the importance of the localization loss. It is typically set to a high value (e.g., 5) to ensure the model prioritizes learning to accurately locate objects.
    \item $\lambda_{\text{noobj}}$ (Lambda No-Object): This hyperparameter reduces the influence of the no-objectness loss. Since most grid cells do not contain any objects, the no-object confidence loss would easily dominate the overall loss. A low value (e.g., 0.5) is used to balance this effect.
\end{itemize}

To transform 2D pixel-level BB classifications into spatially grounded, human-interpretable metrics, the proposed framework extends the edge-comp\\-uted YOLOx detections into 3D metric space. This spatial alignment is essential for a field-deployable DSS, as farmers and robots must appear at the correct coordinates on the robots’ edge maps and on the iVeg map for the farmer; while raw 2D bounding boxes hold little semantic meaning for a grower monitoring fleet operations, 3D coordinates relative to the greenhouse topology provide actionable, transparent insights. To achieve this without streaming high-bandwidth data, the local robotic nodes utilize an optimized 3D projection pipeline inspired by the YOLACT\_ros\_3d\footnote{YOLACT\_3D repository: \url{https://github.com/IntelligentRoboticsLabs/yolact_ros_3d/tree/master}} framework, which maps localized instance representations directly against the registered depth point cloud captured by the Intel RealSense D435 camera. The local detection masks are encoded as optimized bitsets. For a given pixel at coordinates $(x, y)$, where $x$ represents the column and $y$ represents the row, the membership function within the localized region of interest is defined as:

\begin{equation}
\text{belongsToMask}(x, y) = \begin{cases} 
1 & \text{if bit}_{b,n} = 1 \\
0 & \text{otherwise}
\end{cases}
\end{equation}

\noindent where the bit position is computed as:

\begin{equation}
\text{index} = y \cdot w + x
\end{equation}

\begin{equation}
b = \lfloor \text{index} / 8 \rfloor
\end{equation}

\begin{equation}
n = 7 - (\text{index} \mod 8)
\end{equation}

\noindent and $w$ is the mask width, $b$ is the byte index, and $n$ is the bit index within that byte. Protruding greenhouse foliage often fringes the visual boundaries of targets, causing erratic depth readings that degrade human-robot trust. To filter this noise, an adaptive morphological erosion is applied before 3D projection. The kernel area $A_k$ relies on a class-specific erosion factor $e_c \in [0, 100]$:

\begin{equation}
A_k = \max\left(\text{MINKERNEL}, \text{MAXKERNEL} - \frac{e_c \cdot \text{MAXKERNEL}}{100}\right)
\end{equation}

A cross-shaped structuring element ($SE$) of size $k_{\text{size}} = \lfloor \sqrt{A_k} \rfloor \times \lfloor \sqrt{A_k} \rfloor$ iteratively erodes the original mask $M^{(0)}$ via morphological opening until peripheral artifacts detach:

\begin{equation}
M_{\text{eroded}} = \bigcup_{t=1}^{T} \left(M^{(t-1)} \cap \neg\left[(M^{(t-1)} \ominus SE) \oplus SE\right]\right)
\end{equation}

\noindent where $T$ is the total number of iterations. For each eroded pixel ($M_{\text{eroded}}(j, i) = 255$), its 3D coordinate is retrieved. Given the YOLOx 2D center $(x_c, y_c)$, width $w_i$, and height $h_i$, where $j$ and $i$ represent the local row and column indices within the mask, the top-left corner $(x_{\text{tl}}, y_{\text{tl}})$ of the bounding box is computed to map the 2D pixel to the 3D point cloud index:

\begin{equation}
(x_{\text{tl}}, y_{\text{tl}}) = \left(x_c - \frac{w_i}{2}, y_c - \frac{h_i}{2}\right)\end{equation}

\begin{equation}
\text{pc\_index} = (j + y_1) \cdot W + (i + x_1)
\end{equation}

\noindent where $W$ is the total point cloud width. To prevent erratic depth jumps caused by layered crop rows, a point $\mathbf{p}_{\text{pc\_index}} = (p_x, p_y, p_z)^T$ is only accepted if it satisfies a depth discontinuity constraint:

\begin{equation}
| p_{z, \text{pc\_index}} - p_{z, \text{pc\_index} - 1} | \leq \tau_{\text{max}}
\end{equation}

\noindent where $\tau_{\text{max}}$ filters background clipping. This physically validates the semantic confidence $C_i$ of the YOLOx model. The 3D bounding box $\mathbf{BB}_{3D}$ is formed using the filtered set of valid points $\mathcal{V}$:

\begin{equation}
\mathbf{BB}_{3D} = \begin{bmatrix}
x_{\min} & x_{\max} \\
y_{\min} & y_{\max} \\
z_{\min} & z_{\max}
\end{bmatrix}
\end{equation}

\noindent where $x_{\min}$ and $x_{\max}$ define the spatial boundaries along the X-axis, calculated as $x_{\min} = \min_{k \in \mathcal{V}} p_{x,k}$ and $x_{\max} = \max_{k \in \mathcal{V}} p_{x,k}$. Here, $p_{x,k}$ denotes the X-coordinate of the $k$-th 3D point belonging to the filtered vertex set $\mathcal{V}$. The boundary pairs for the Y and Z axes ($y_{\min}, y_{\max}$ and $z_{\min}, z_{\max}$) are derived analogously from their respective spatial components ($p_{y,k}$ and $p_{z,k}$). To optimize edge processing, only detections meeting the confidence threshold ($C_i \geq p_{\text{min}}$) are computed. The final spatial centroid $\mathbf{c}_{3D}$ and dimensions $\mathbf{d}_{3D}$ are:

\begin{equation}
\mathbf{c}_{3D} = \begin{pmatrix} \frac{x_{\max} + x_{\min}}{2} \\ \frac{y_{\max} + y_{\min}}{2} \\ \frac{z_{\max} + z_{\min}}{2} \end{pmatrix}, \quad \mathbf{d}_{3D} = \begin{pmatrix} x_{\max} - x_{\min} \\ y_{\max} - y_{\min} \\ z_{\max} - z_{\min} \end{pmatrix}
\end{equation}

Finally, coordinates are transformed from the robot's egocentric view to a unified global map using ROS 2 \texttt{TF2}, allowing the DSS to present a coherent world model:

\begin{equation}
\mathbf{P}_{\text{work}} = \mathbf{T}_{\text{work}}^{\text{sensor}} \cdot \mathbf{P}_{\text{sensor}}
\end{equation}

This information is updated for every frame in the \texttt{/yolact\_coordinate} topic in ROS 2. Consequently, the topic contains the information shown in Table \ref{tab:object_parameters}

\begin{table}
\centering
\caption{Description of the detected object parameters.}
\label{tab:object_parameters}
\begin{tabular}{|l|p{10cm}|}
\hline
\textbf{Parameter} & \textbf{Description} \\ \hline
\textit{object\_name} & Object name. \\ \hline
\textit{probability} & Probability of certainty. \\ \hline
\textit{xmin} & X coordinate in meters of the left upper corner of the bounding box. \\ \hline
\textit{xmax} & X coordinate in meters of the right lower corner of the bounding box. \\ \hline
\textit{ymin} & Y coordinate in meters of the left upper corner of the bounding box. \\ \hline
\textit{ymax} & Y coordinate in meters of the right lower corner of the bounding box. \\ \hline
\textit{zmin} & Z coordinate in meters of the nearest pixel of the bounding box. \\ \hline
\textit{zmax} & Z coordinate in meters of the furthest pixel of the bounding box. \\ \hline
\end{tabular}
\end{table}

In conclusion, the YOLO framework algorithm follows these steps:

\begin{enumerate}
    \item The Intel RealSense D435 camera captures a frame structured as a pixel matrix.
    \item YOLOx processes this matrix and locates an object (for example, the ''farmer`` class). The network generates a 2D bounding box defined by its centre, width and height in pixels.
    \item From there, YOLACT generates a binary mask (a bitmap) cropped exactly to the size of that bounding box. $x$ and $y$ are the iterative indices used by the loop in your code to traverse, pixel by pixel, that local mask. 
    \item If the pixel at position $(x, y)$ has a bit set to 1 (i.e., belongsToMask(x, y) = 1), this means that this particular pixel is part of the farmer’s or robot’s body and not part of the background.
\end{enumerate}

\subsubsection{MQTT}
\begin{figure}[!ht]
    \centering
    \includegraphics[width=0.7\textwidth]{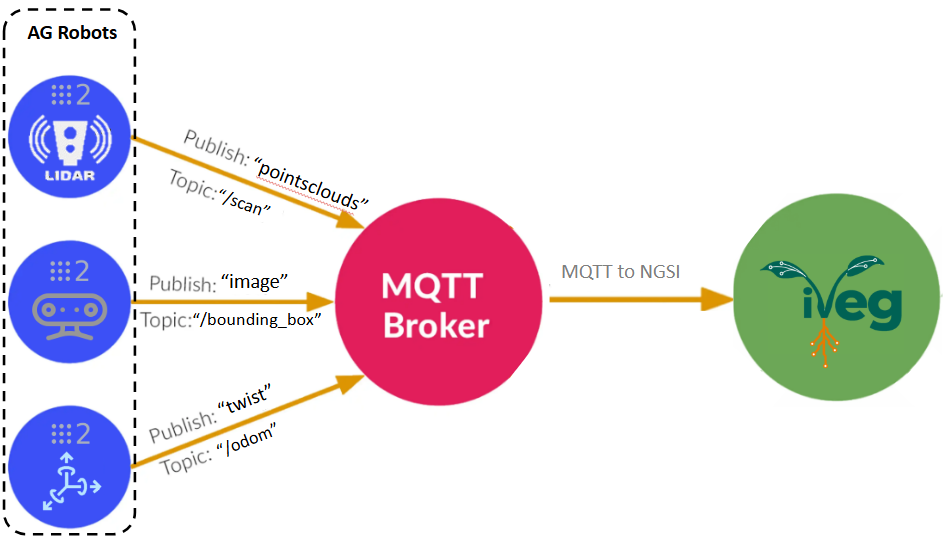}
    \caption{MQTT communication}
    \label{fig:ros2mqtt}
\end{figure}

MQTT (Message Queuing Telemetry Transport) is a lightweight communication protocol designed for IoT and M2M applications. Its publication and subscription-based architecture allows information to be exchanged efficiently through a centralized model managed by a broker, which organizes messages into topics accessible to publishers and subscribers. One of its most outstanding features is its optimization for resource-constrained devices and low-capacity networks, thanks to a fixed header size of only 2 bytes. MQTT includes three Quality of Service (QoS) levels, which guarantee the delivery of messages with varying degrees of reliability. In this design, the AGI and AGII robots communicate by sending information from the relevant ROS 2 topics, as shown in Figure \ref{fig:ros2mqtt}.

\subsubsection{FIWARE}

FIWARE is a framework developed as part of a European initiative to drive the adoption of IoT platforms and foster sustainability in smart applications \citep{FIWARE}. This modular system is based on a set of Generic Enablers (GEs) that simplify the development of IoT solutions through standardized functionalities such as contextual data management, open APIs, and information processing.

A key element of FIWARE is the Orion Context Broker (OCB), which acts as the core of any solution based on this framework, managing contextual data in real time using the OMA NGSI model \citep{kovacs2016standards}. IoT Agents play a crucial role in translating information from IoT devices into the NGSI standard format, supporting protocols such as MQTT \citep{mqtt}, CoAP \citep{Shelby2014}, and OPC UA \citep{OPCUA}. This interoperability makes FIWARE especially relevant for applications that require efficient integration and scalability in heterogeneous environments, such as smart greenhouses.

\subsubsection{iVeg DSS Interface}

iVeg serves as the user-facing interface within this architecture. Built on top of FIWARE, it transforms the standardized NGSI data streams into an interactive, web-based situational awareness platform. Instead of merely displaying raw robot coordinates, iVeg contextualises the decision-making process for the greenhouse manager. It provides real-time, synchronized representations of the environment, visualizing the exact locations of the fleet alongside the 3D bounding boxes of detected human operators or obstacles, displaying the classes detected on the robots’ local maps and on the website set up for farmers. By exposing the fleet's intent, dynamic costmaps, and decision certainty thresholds, iVeg bridges the gap between robotic automation and human oversight, fostering trust in the system's reliability. Its modular architecture facilitates the incorporation of new devices, services, and analytical tools, making iVeg suitable for research and experimentation in smart agriculture and robotic applications.

\subsubsection{Sensor Fusion and Decision Algorithm}

The proposed decision-making algorithm integrates heterogeneous sensor data for real-time environmental interpretation and cloud-based synchronisation via the Internet of Things (IoT). Each robotic unit is equipped with a HITTSON onboard computer acting as the central processing hub, interfaced with an RGB-D camera, LiDAR, IMU, and motor encoders to continuously evaluate the operational workspace. This unit identifies human operators and robotic peers (AGI/AGII) to facilitate data dissemination across the local network infrastructure, all within the edge computing environment. The perception pipeline employs the YOLOx framework—specifically fine-tuned for this application—for real-time inference on the incoming video stream, extracting class identifiers and confidence scores for transmission via the MQTT protocol to a FIWARE-based cloud platform. Furthermore, a spatial localisation algorithm processes the YOLOx bounding boxes to determine the 3D positions of the detected objects relative to the camera's optical centre; this spatial metadata is also forwarded to the iVeg DDS. Within this architecture, iVeg filters these detections, ensuring that only entities that meet a predefined probability threshold are classified as confirmed agents, thereby providing the entire fleet with the precise locations of nearby farmers and other classes. To enable distributed situational awareness, each robot transmits odometry data derived from LiDAR and wheel encoders to determine its precise global pose. This information is shared across the fleet to update local costmaps with the positions of dynamic obstacles in real time, allowing the system to track both human operators and robotic peers simultaneously. 

The DDS enforces safety through a hierarchical rule-based logic that prioritises human safety. Specifically, when YOLOx detects a farmer, the system broadcasts the 3D coordinates to a dedicated node on each robot that continuously monitors the MQTT data stream. If incoming data confirms human presence within the greenhouse, the system generates a 2-meter-radius "warning zone" around each operator. Within this zone, robots are mandated to reduce their operational speed by half. In the absence of human presence, the system manages multi-robot interactions by evaluating the task status of detected peers. When a robot identifies another unit, its location is shared via MQTT/FIWARE, and a similar warning radius is established to mitigate collision risk. These velocity adjustments are implemented directly within the motion control node, which applies a 0.5 scalar multiplier to the \texttt{cmd\_vel} topic, effectively reducing the voltage supplied to the actuators. Ultimately, this multi-sensor fusion approach ensures that safety constraints override standard navigation commands, thereby enhancing the reliability of autonomous operations in the dynamic, unstructured conditions of protected cultivation.

\section{Robot-robot and human-robot collaboration architecture} \label{sec: Robot}

The AgroConnect greenhouse is equipped with an advanced IoT-enabled architecture designed to facilitate efficient data management and interoperability. The architecture integrates multiple layers—perception, processing, and application—ensuring seamless collaboration between heterogeneous components such as sensors, actuators, and cloud services. Although this architecture was presented in a basic version in a previous paper \citep{munoz2020new}, the current implementation has evolved to incorporate advanced robotic capabilities, enabling autonomous operations and improved data exchange. This section describes the proposed architecture, beginning with an overview of the existing layers within the greenhouse and the communication methodology between robots.

\subsection{IoT-based architecture in the AgroConnect greenhouse}

\begin{enumerate}
    \item Perception Layer: The perception layer represents the foundation of the architecture, consisting of various physical components deployed in the greenhouse. This includes:

    \begin{itemize}
        \item Sensors and IoT devices: Used for monitoring key environmental variables such as temperature, humidity, and CO2 levels.
        \item Robots (AGI and AGII): Integrated for cooperative tasks such as navigation, data collection, etc.
        \item Other equipment: Components such as thermal panels, fertigation systems, and reverse osmosis devices contribute to the efficient management of resources.
    \end{itemize}

    \item Data Storage and Processing Layer:  This layer is central to managing and processing data generated by the perception layer:

    \begin{itemize}
        \item FIWARE Enablers: Components such as the Orion Context Broker, IoT Agents, and Cygnus handle real-time contextual data and ensure interoperability using the OMA NGSI standard.
        \item MQTT Broker: Facilitates lightweight communication between sensors, robots, and cloud services.
        \item ETL Processes: Support the extraction, transformation, and loading of data into dedicated databases (e.g., MongoDB, PostgreSQL) for further analysis
    \end{itemize}

    \item Application Layer:  The application layer provides interfaces and services for decision-making and visualization:

    \begin{itemize}
        \item Dashboards: Enable real-time monitoring of greenhouse conditions and robot operations.
        \item REST APIs: Offer external access to processed data for integration with external systems or services.
        \item Digital Twins and Advanced Analytics: Allow simulation and predictive modeling to optimize greenhouse operations.
    \end{itemize}
\end{enumerate}

\begin{figure}[!ht]
	\centering
\includegraphics[ width=0.8\textwidth]{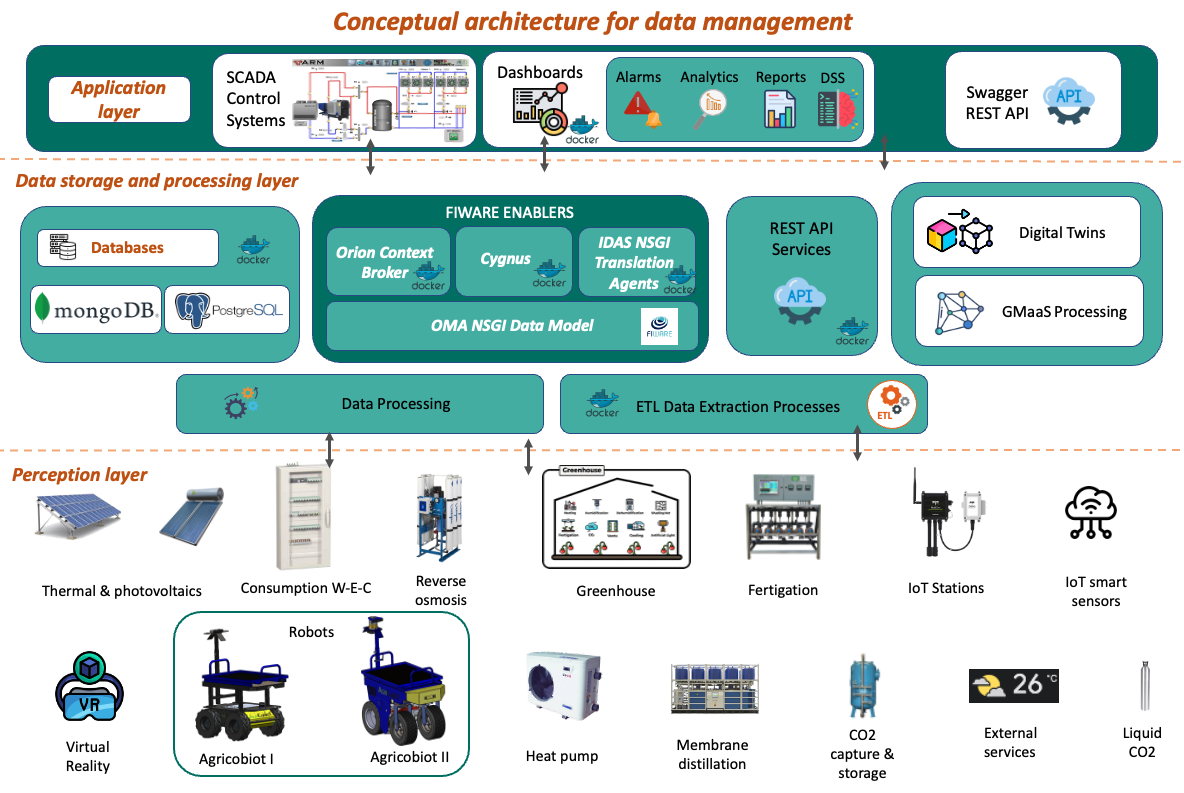}
	\caption{Conceptual architecture for data management.}
	\label{fig:arq}
\end{figure}

This layered architecture (see Figure \ref{fig:arq}) not only ensures efficient data processing and interoperability but also supports real-world operations through seamless integration of sensors, robots, and cloud services. The following section illustrates specific examples of how this architecture enables practical workflows in the AgroConnect greenhouse.

\subsection{Cooperative and Collaborative Robot Logic}

This section describes the workflow that enables an example of cooperation between the AGI and AGII robots within the greenhouse on the iVeg platform (see Figure \ref{fig:iVeg2}) when one robot detects another via its sensors or by sending a command from the control panel. Furthermore, when the robots detect a farmer, they also send them the necessary information so that both robots can take the appropriate actions, adapting their behaviour to the presence of humans. This communication is managed through the publication and processing of data via the MQTT broker, conversion to NGSI in the IoT Agent, and the generation of commands in the Orion Context Broker. The interaction and communication with the dashboard are described below.

\subsubsection{Robot-Robot Bidirectional Bridge}

To ensure decentralized safety, the fleet maintains a continuous, low-latency communication mesh by ROS 2 Humble and a custom-developed Bridge Node. The core of this integration is a custom-developed Bridge Node, which functions as a bidirectional gateway between the robotic middleware and the cloud infrastructure. Internally, ROS 2 utilizes a Publish-Subscribe pattern where vision nodes (YOLO) and the MVSim simulator (for virtual environments) or the physical robot (for real-world trials) provide asynchronous data streams.

\begin{figure}[!ht]
    \centering
    \includegraphics[width=\textwidth]{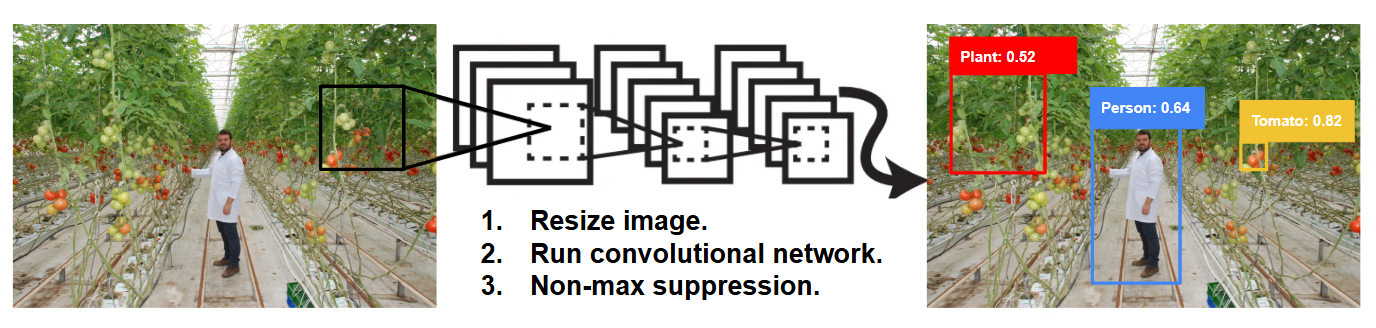}
    \caption{YOLO process recognition}
    \label{fig:YOLO2}
\end{figure}

\begin{figure}[!ht]
    \centering
    \includegraphics[width=0.75\textwidth]{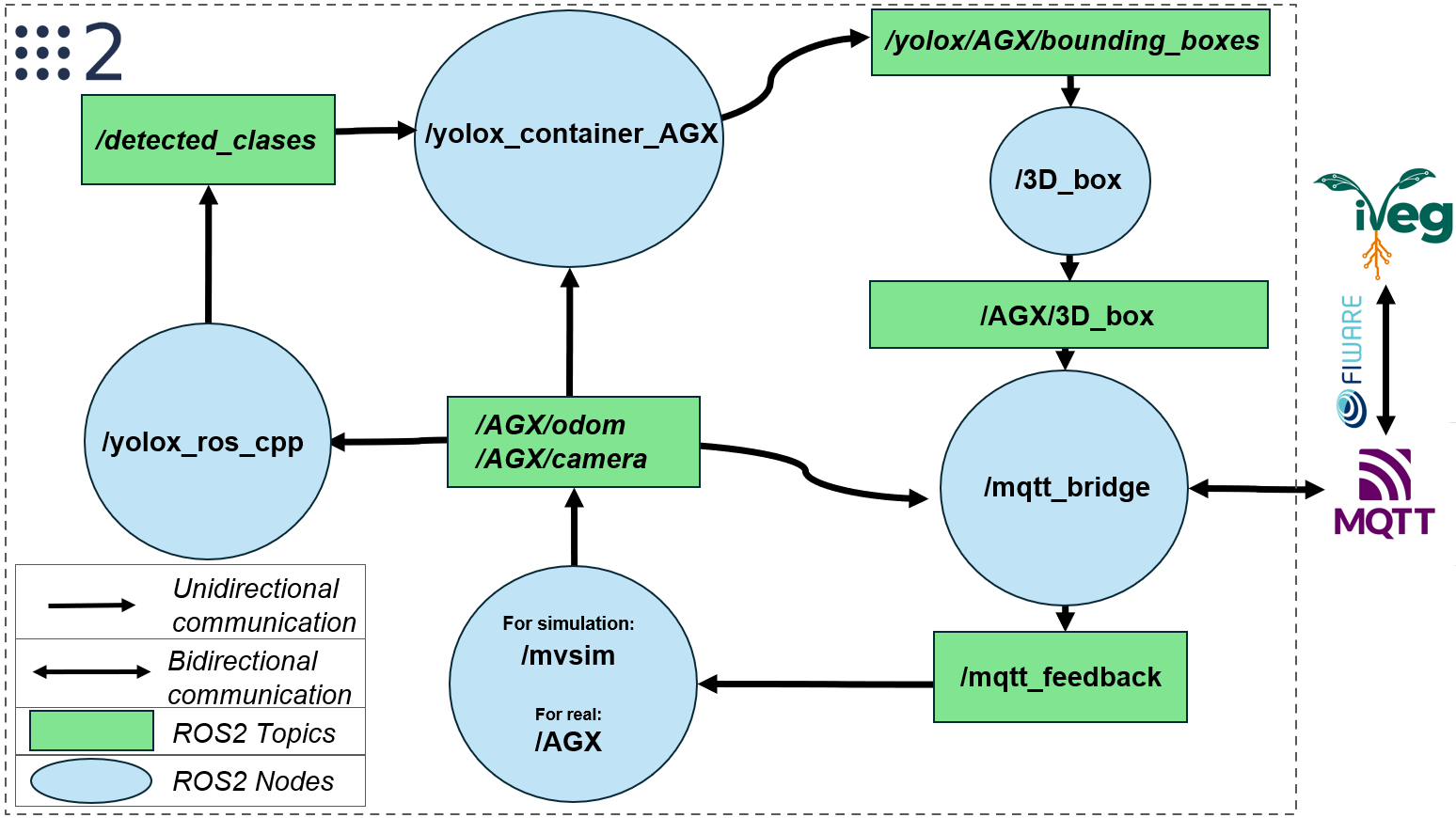}
    \caption{Sample code for the ROS 2 to MQTT bridge, FIWARE and iVeg, where X refers to the AGI or AGII robot}
    \label{fig:ROSexpl}
\end{figure}

The Bridge Node serializes high-level metadata at 30 Hz, packing the robot's states, linear/angular velocities, and active 3D YOLO \texttt{BoundingBoxes}. Upon reception, the system executes a \textit{JSON serialization} process: rather than forwarding bandwidth-heavy raw images, it filters classification metadata—such as Class ID and probability—and encapsulates it into a lightweight MQTT payload (see Algorithm \ref{algo:1} and Figure \ref{fig:YOLO2}). Simultaneously, the node processes ground-truth odometry and transformed LiDAR scan vectors to synthesize a shared, cross-fleet costmap. This architecture ensures that every autonomous unit maintains persistent, identical knowledge of its peers' physical boundaries without overwhelming the greenhouse's wireless network bandwidth (see Figure \ref{fig:ROSexpl}).

A specialized background node, \texttt{mqtt\_feedback}, continuously queries the cloud broker to retrieve the real-time spatial trajectories of neighboring agents. Concurrently, the deterministic decision-making logic (detailed in Section \ref{sec_iVeg}) tracks these incoming classification labels and confidence metrics (see Algorithm \ref{algo:2}). 

Whenever an agent registers a person or a peer, a fleet-wide geospatial alert is broadcasted. Upon receiving this message, the neighboring robot computes an immediate proximity analysis relative to its own kinematic path. If the trajectory intersects with the 2-meter safety zone of a farmer or the safety buffer of a peer, the recipient robot initiates a preemptive velocity reduction (see Algorithm \ref{algo:3}). This mechanism acts as an extended safety horizon: it grants the robot spatial awareness of human presence well beyond its immediate line-of-sight, overrules active navigation commands to force a predictable deceleration, and exposes this entire safety state to the iVeg interface for complete human validation. Finally, it updates the map in RViz and displays the class detected by the other robot to define the safety radius. At the same time, iVeg updates its map on the dashboard so that the farmer can see where the other farmers are.

\begin{algorithm}[!ht]
\caption{AI-edge computing and data transmission to iVeg via MQTT}
\label{algo:1}
\begin{algorithmic}[1]
\Require ROS 2 Topics: $\mathtt{/odom}$, $\mathtt{/scan}$, $\mathtt{/yolox/bounding\_box}$, $\mathtt{/yolact\_coordinate}$
\Ensure MQTT Topic: $\mathtt{/\{apikey\}/\{robot\_id\}/attrs}$

\Procedure{OnSensorDataReceived}{$bbox, bbox_{3D}, odom, scan$}
    \State $\mathbf{x}_{\text{robot}} \gets odom.\text{pose}.\text{position}$ \Comment{Extract egocentric position}
    \State $\theta_{\text{robot}} \gets \Call{GetHeading}{odom.\text{pose}.\text{orientation}}$
    
    \State $\mathbf{P}_{\text{work}}, c, p_i(c) \gets \Call{SpatialGrounding}{bbox_{3D}, bbox}$ \Comment{Transform to world frame}
    
    \State $payload.\text{class} \gets c$
    \State $payload.\text{conf} \gets p_i(c)$
    \State $payload.\text{position\_3D} \gets \mathbf{P}_{\text{work}}$
    \State $payload.\text{pose} \gets odom.\text{pose}$
    \State $payload.\text{laser\_scan} \gets \Call{DownsampleScan}{scan, \text{resolution}=30\text{Hz}}$
    \State $payload.\text{timestamp} \gets t_{\text{now}}$
    
    \State $\mathtt{topic} \gets \text{"/"} + \mathit{apikey} + \text{"/"} + \mathit{robot\_id} + \text{"/attrs"}$
    \State \Call{MqttPublish}{\texttt{topic}, payload, \text{QoS}=0}
\EndProcedure
\end{algorithmic}
\end{algorithm}

\begin{algorithm}[!ht]
\caption{Cloud-Based DSS Processing}
\label{algo:2}
\begin{algorithmic}[1]
\Require Context Entities tracked in Orion Context Broker via IoT Agent MQTT Bridge
\Ensure MQTT Topic Feedback: $\mathtt{/\{robot\_id\}/mqtt\_feedback}$

\Loop
    \State $data \gets \Call{QueryOrionContext}{\mathit{robot\_id}\text{/attrs}}$ \Comment{Synchronous polling from OCB}
    
    \State $\Call{Up.SemanticDashboard}{data.\text{position\_3D}, data.\text{class}, data.\text{pose}}$ 
    \Comment{Render UI}
    
    \If{$data.\text{conf} \ge 0.80$}
        \If{$data.\text{class} \in \{\text{'person'}, \text{'robot'}\}$}
            \State $alert \gets \Call{}{data.\text{position\_3D}, data.\text{class}, data.\text{conf}, t_{\text{now}}}$
        \Else
            \State $alert \gets \Call{}{t_{\text{now}}}$
        \EndIf
    \EndIf
    
    \State $robot\_pose \gets \Call{}{data.\text{pose}, data.\text{laser\_scan}, t_{\text{now}}}$
    \State \Call{MqttPublish}{\text{"/"} + \text{robot\_id} + \text{"/mqtt\_feedback"}, [alert, robot\_pose], \text{QoS}=1}
\EndLoop
\end{algorithmic}
\end{algorithm}

\begin{algorithm}[!ht]
\caption{Local decision based on information from the DSS}
\label{algo:3}
\begin{algorithmic}[1]
\State \textbf{Input:} MQTT Topic: \texttt{/\{peer\}/mqtt\_feedback}, ROS 2 Topic: \texttt{/cmd\_vel\_input}, \texttt{/obstacle\_DSS}
\State \textbf{Output:} Modulated ROS 2 Topic: \texttt{/cmd\_vel\_output}

\Procedure{OnMqttFeedback}{$data$}
    \Loop
        \State $iden\_farmer\_robot \gets$ \Call{Decode}{$data.alert$}
        \State $farmer\_pos \gets$ \Call{Decode}{$data.3Dposition$}
        \State $ag\_robot\_pos \gets$ \Call{Decode}{$data.pose$}
        \If{$iden\_farmer ==$ TRUE}
            \State update local\_map $\gets iden\_farmer$ \Comment{set farmer/robot pos. in local map and generate security circle}
            \If{$farmer\_pos =<$ \texttt{AG/odom}}
                \State $V_{linear} \gets V_{input} \times 0.5$
                \State \Call{Publish}{"cmd\_vel", $V_{linear}$}
                \State \textbf{Log:} "Speed reduced: Within the farmer's safety zone"
            \EndIf
        \EndIf
        \If{$ag\_robot\_pos =<$ \texttt{AG/odom}}
            \State $V_{linear} \gets V_{input} \times 0.5$
            \State \Call{Publish}{"cmd\_vel", $V_{linear}$}
            \State \textbf{Log:} "Speed reduced: Within the robot's safety zone"
        \EndIf
    \EndLoop
\EndProcedure
\end{algorithmic}
\end{algorithm}

\subsection{Communication workflow between robots by the iVeg Dashboard} \label{sec_iVeg}

\begin{figure}[!ht]
    \centering
    \includegraphics[width=0.9\textwidth]{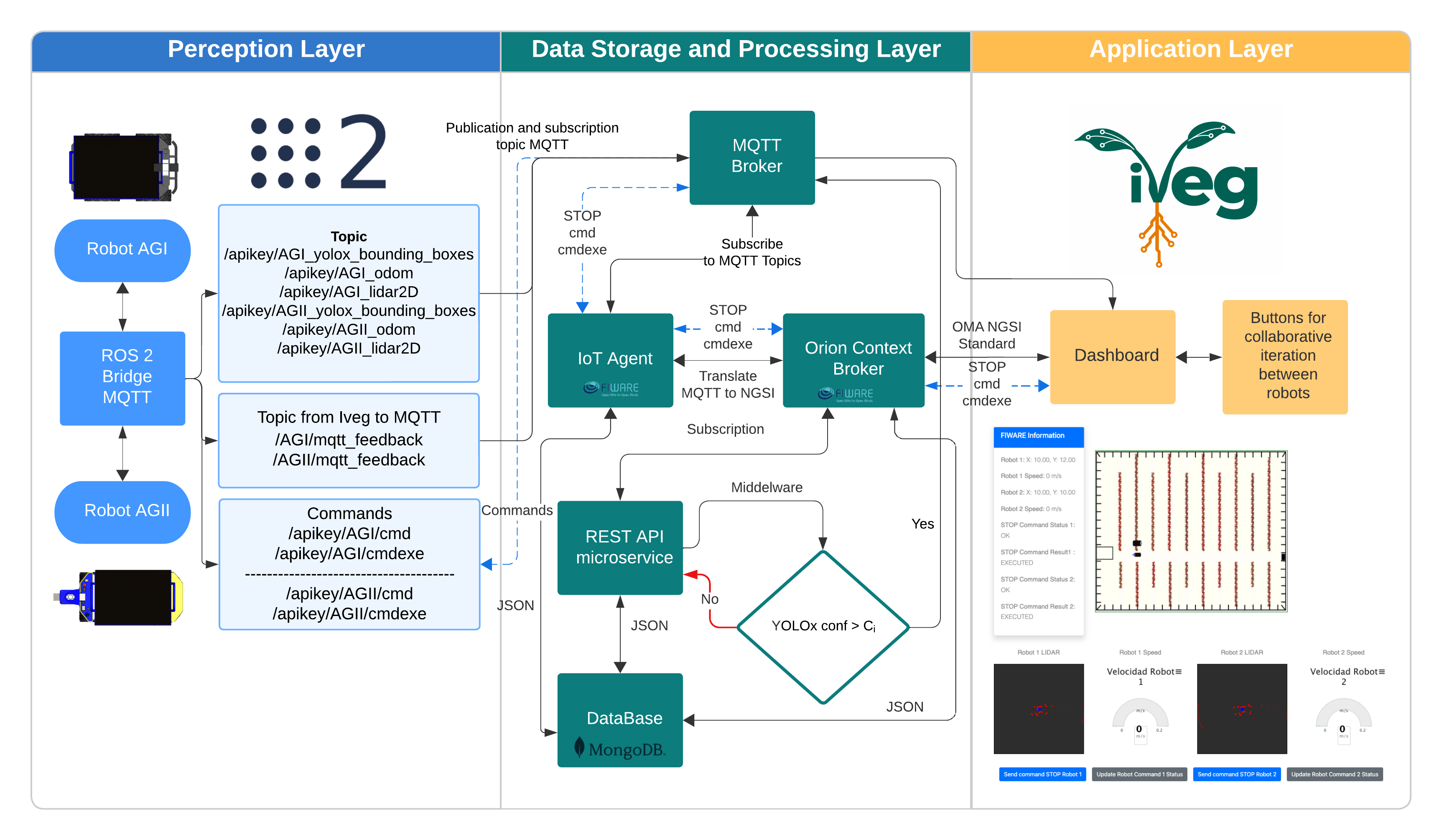}
    \caption{Communication and cooperation flow between the robots and the iVeg platform.}
    \label{fig:iVeg2}
\end{figure}

This section details the comprehensive semantic workflow designed to operationalize real-time, explainable monitoring and multi-agent cooperation between the AGI and AGII units, as well as their context-aware interaction with human operators via the iVeg platform (see Figure \ref{fig:iVeg2}). By leveraging an IoT-standardized layered architecture anchored on MQTT and OMA NGSI, the framework transforms raw, uninterpretable edge-computed sensor data into high-level, human-centric actionable insights for transparent fleet management.

\begin{enumerate}
    \item \textit{Perception Layer - Data Generation and Publication (Onboard):}
    Each robot gathers critical navigation and environmental data through its sensor suite, including LiDAR, RGB-D cameras and odometry. The \textit{ROS 2 Bridge MQTT} component acts as an intermediary, converting internal ROS 2 messages into lightweight MQTT topics. This information is published to the \textit{MQTT Broker} under the following topics, following the IoT Agent format \texttt{/\{apikey\}/\{device\}/attrs}:
    \begin{itemize}
        \item \texttt{/\{apikey\}/AGI\_odom} and \texttt{/\{apikey\}/AGII\_odom}: Robot position ($x, y$), orientation ($yaw$), and velocities ($v,\omega$).
        \item \texttt{/\{apikey\}/AGI\_lidar2D} and \texttt{/\{apikey\}/AGII\_lidar2D}: LiDAR point clouds for localization.
        \item \texttt{/\{apikey\}/AGI\_yolox\_bounding\_boxes} and \texttt{/\{apikey\}/AGII\_\\yolox\_bounding\_boxes}: YOLOx detection metadata (class, confidence, bounding box, 3D position).
    \end{itemize}

    \item \textit{Data Storage and Processing Layer - Standardization and Cloud Integration:} 
    To ensure that the fleet's cognitive state is interoperable and open to external auditing, the telemetry is routed through a cloud-based standardization pipeline:

    \begin{itemize}
        \item Reception and Translation via IoT Agent: The IoT Agent subscribes to the relevant MQTT topics and functions as a gateway that translates incoming data into the OMA NGSI format. This ensures data interoperability and seamless integration into the FIWARE ecosystem. For instance, data from \texttt{/\{apikey\}/AGII\_odom\\/attrs} is transformed into an NGSI entity with \texttt{base\_pose\_groun\\d\_truth} attributes, while \texttt{/\{apikey\}/AGII\_lidar2D/attrs} maps to \texttt{lidar\_points} and \texttt{/\{apikey\}/AGII\_yolox \_bounding\_boxes\\/attrs} maps to \texttt{bounding\_boxes}.
    
        \item Integration with Orion Context Broker: The IoT Agent forwards the translated data to the Orion Context Broker, where it is stored as FIWARE entities available for real-time queries. In Listing 1 an example of a unified robot state entity and its standardized attributes is shown.

        \item MongoDB Database: The Orion Context Broker shares a MongoDB database with the IoT Agent and the REST API for historical data persistence, enabling longitudinal analysis and trend identification.
    \end{itemize}
    
    \begin{lstlisting}[caption=Structure of the NGSI-v2 Entity for the Robot’s Status in Orion Context Broker, basicstyle=\small\ttfamily, columns=fullflexible, xleftmargin=1em] 
  {
    "id": "AGX",
    "type": "Robot",
    "base_pose_ground_truth": {
        "type": "object",
        "value": {
            "x": 14.134062767,
            "y": 11.46727562,
            "yaw": -0.386157338,
            "linear_velocity": 0.999,
            "angular_velocity": 0.34
        }
    },
    "bounding_boxes": {
        "type": "object",
        "value": {
            "class_id": "farmer",
            "probability": 0.945,
            "position": { 
              "xmin": 0.4506256580352783
              "ymin": -0.3164764642715454
              "xmax": 0.7936256527900696
              "ymax": 0.11368180811405182
              "zmin": -0.25958430767059326
              "zmax": 0.10506562888622284
            }
        }
    },
    "lidar_points": {
        "type": "array",
        "value": [
            5.3208699226379395,
            5.330774307250977,
            5.321296691894531,
            ...
        ]
    },
    "metadata": {
        "TimeInstant": {
            "type": "DateTime",
            "value": "2026-04-26T10:21:14.660Z"
        }
    }
}
    \end{lstlisting}

    \item \textit{Application Layer - iVeg Dashboard Visualization:} 
    
    The iVeg platform acts as the bridge between machine logic and human supervision, translating complex digital twin data streams into global situational awareness. This layer operationalizes the transparency goals through four core capabilities:
    
    \begin{itemize}
        \item \textit{Reception and Translation via IoT Agent:} The primary contribution to the collaborative workflow is the deterministic propagation of risk awareness. When the iVeg DSS processes an entity classification matching a ''person`` or ''robot`` with confidence score $C_i \ge 0.80$, it automatically triggers a fleet-wide feedback loop. The system formats and publishes this alert metadata (\texttt{data.position}, \texttt{data.class}, \texttt{data.conf}, \texttt{timestamp}) to the corresponding feedback topic (\texttt{/AGI/mqtt\_feedback} or \texttt{/AGII/mqtt\_feedback}), mandating an immediate, proactive 50\% kinematic speed override on any unit operating within that zone.
        
        \textit{Updating the dashboard’s semantic map:} Instead of displaying abstract lists of coordinates, the interface generates an interactive semantic map in real time. Following confirmation by the DSS, the user interface automatically projects a spatial marker that dynamically traces a 2-metre safety radius around the human operator, both on the fleet’s shared-cost maps and on the manager’s control panel. 
        
        \item \textit{Human-in-the-Loop Kinematic Override:} To preserve human agency over autonomous behaviors, the UI features explicit manual control mechanisms. Operators can instantly issue an emergency \texttt{STOP} command. This instruction is routed via NGSI-v2 down to the IoT Agent, which publishes a high-priority payload to the robot's command topic (\texttt{/\{apikey\}/AGI/cmd} or \texttt{/\{apikey\}/AGII/\\cmd}), forcing an immediate mechanical shutdown that completely overrides any ongoing autonomous navigation task.

        \item MongoDB Database: The Orion Context Broker shares a MongoDB database with the IoT Agent and the REST API for historical data persistence, enabling longitudinal analysis and trend identification.

    \end{itemize}
\end{enumerate}

\section{Results and Discussion} \label{sec: result}

This section describes the simulations and experiments designed to validate the proposed approach, as well as the results obtained.

\subsection{Validation in the Simulated Greenhouse} \label{sec:intri}

For this study, a preliminary simulation was designed and executed within a 3D greenhouse model previously established in \citep{canadas2026ros2}, utilizing real-world data collected from the \citep{canadas2024greenbot} dataset. The simulation environment was developed using the MVSim simulator \citep{blanco2023multivehicle}, which is particularly suited for these scenarios due to its high-fidelity physical modeling. MVSim provides realistic friction models, a critical factor when addressing the complexities of greenhouse environments characterized by uneven soil and loose earth surfaces.

A total of five simulations were conducted across four distinct scenarios: i) farmer detection from various perspectives; ii) AGX robot identification from multiple angles; iii) simultaneous appearance of a farmer and AGX unit within the same frame; and iv) farmer and AGX unit detection under partial occlusions within the field of view. Out of all trials, 18 were successfully completed, yielding an overall success rate of 92\%. The details of the simulation and the results obtained are set out below.

\subsubsection{Simulation setup}

This section describes the simulation setup, which will be used as a model for real-world tests. 

\begin{itemize}
    \item \textit{3D model of the greenhouse}
\end{itemize}

For this work, a 20 × 20 m virtual greenhouse model was constructed, consisting of five corridors, each bordered on both sides by tomato plants and separated by 4 m, thereby defining the robot’s navigation lanes. This model has previously been used in the study \citep{Canadasifac2026}, supporting its application to the real-world model. The 3D model closely replicates a real pear-tomato crop grown in a hydroponic system and the characteristic geometry of a Mediterranean greenhouse. This realistic representation provides a reliable testbed for validating navigation algorithms previously applied in related research \citep{canadas2024pid, Canadas_robotdesing}. The model is shown in Figure \ref{fig:mvsim} and, to adapt it to the real-world environment, four realistic models of farmers, stacked crates of tomatoes and two transport trolleys have been implemented; these models correspond to those found in the actual AgroConnect greenhouse.

\begin{itemize}
    \item \textit{AGI 3D model}
\end{itemize}

The AGI robotic asset simulates an autonomous scouting platform. All onboard sensory data streams are structurally referenced to its egocentric $\mathtt{base\_link}$ coordinate frame. The payload comprises a 3D LiDAR (Velodyne VLP16) dedicated to planar mapping and a forward-facing RGB-D camera (Intel RealSense D435) rigid-mounted at coordinates (0.4, 0.0, 0.6) m to optimize the optical field of view. The camera’s factory-calibrated intrinsic parameter matrices are integrated directly into the local ROS 2 projection nodes, enabling the instantaneous translation of 2D bounding boxes into metric 3D coordinates relative to the optical center. The RGB-D sensor provides synchronised colour and depth information, enabling the direct estimation of the distance to detected objects from the bounding box in the YOLOACT algoritm. 

\begin{itemize}
    \item \textit{AGII 3D model}
\end{itemize}

The AGII asset represents a heavy-duty crop transport platform operating under loaded structural conditions. It is equipped with a co-aligned 3D LiDAR (Ouster OS0) and a stereo vision system (Point Grey Bumblebee2) mounted at a baseline height of (0.35, 0.0, 0.6) m. The stereo system extracts spatial depth via localized disparity matching between the synchronized left and right image frames, providing a secondary, independent pipeline for 2D-to-3D projection verification. The intrinsic parameters used are detailed in \citep{canadas2024greenbot}. The fixed baseline of the stereo system enables reliable depth estimation, allowing the 3D position of detected objects to be computed relative to the robot. 

\subsubsection{Training YOLO for robot recognition}

Regarding the robot detection process, the YOLOx model was trained for 25 epochs with an input resolution of $224 \times 224$ pixels using a Pascal VOC format. The dataset comprised 2,200 images (1,100 of each AGI and AGII robot, including 3D and real-world models ) covering various perspectives and occlusion scenarios within the greenhouse environment. Such training was not required for the farmer class, as a pre-trained human recognition model was included. The dataset was partitioned into \texttt{pascal/train} and \texttt{pascal/valid} directories. Training was executed via the command \texttt{model.train(data='data.yaml', epochs=25, imgsz=224, batch\_size=8, plots=True)} and subsequently exported to the OpenVINO format using \texttt{model.export(format="openvino")}. The training phase was conducted on a local workstation equipped with an NVIDIA RTX 4060 GPU (8 GB), an Intel Core i7-13400 CPU, 32 GB of Kingston DDR4 RAM (3200 MHz), and a 1 TB Kingston NV2 NVMe SSD. This hardware configuration provided sufficient computational throughput and memory bandwidth for real-time image processing, Convolutional Neural Network (CNN) training, and validation tasks, ensuring stable and efficient model convergence.

The system is capable of inter-agent detection with an average confidence probability of 85\%. Figures \ref{fig:YOLOI} and \ref{fig:yolo_metrics_sim} illustrate the identification process of the AGI and AGII robots during simulation trials, as well as the identification of the farmer. Furthermore, the YOLOx performance metrics are summarized in Table \ref{tab:yolo_metrics_sim} using the \texttt{eval.py} file provided by the algorithm. The results demonstrate that the model operates with high precision, reliably performing recognition tasks for both robotic agents and human operators.

The network achieves a global mean precision of 89.90\% and an average precision (AP@0.5) of 0.91. Crucially, the system exhibits an exceptionally low False Negative (FN) profile for the ''Farmer`` class ($\text{FN} = 1$, $\text{Recall} = 97.67\%$). Minimizing false negatives is paramount for human-centric robotics; it ensures that the system rarely misses a human operator under challenging canopy conditions, thereby preventing unexpected or un-auditable movements that could severely degrade grower trust. Consequently, certain real-world challenges—such as variable lighting, motion blur, and partial occlusions—which may occasionally lead to classifications, were not fully encountered. To address these limitations, the following section details the deployment of the robots in a physical greenhouse environment to evaluate their robustness under real-world operational constraints.

\subsubsection{Real-time updating of the semantic map via in simulation}

This section examines the multi-robot communication architecture based on a DSS at iVeg in simulation, detailing how the map is updated in real time locally on the robots and how the map on the iVeg dashboard is updated to facilitate interaction with the farmer.

\begin{itemize}
    \item \textit{Communication and update of the local robot semantic map}

    To evaluate the proposed framework's capability to mitigate spatial blind spots and extend the perceptual horizon, a collaborative non-line-of-sight (NLOS) validation scenario was executed. A critical challenge in dense greenhouse cultivation is the severe occlusion of human operators or mobile assets caused by heavy foliage canopies and structural crop rows. To replicate this real-world operational constraint, the AGII logistic robot was initialized in a navigation corridor completely devoid of a direct line of sight to either the AGI scouting platform or the human operator. Under traditional, isolated control architectures, AGII would remain entirely unaware of peripheral hazards until an immediate, near-collision event occurred. Within the proposed transparency framework, the cloud-based Decision Support System (DSS) functions as a decentralized cognitive bridge. As the AGI platform navigates along the central greenhouse corridor, its edge-computing pipeline continuously scans the environment. The moment the onboard YOLOx inference engine detects the human operator beneath dense canopy occlusions (see Fig.~\ref{fig:YOLO_simu2}), the spatial grounding node estimates the target's absolute metric coordinates within the global frame. This high-level semantic metadata is immediately encapsulated and streamed to the iVeg DSS via the MQTT transparency channel.
    
    The Orion Context Broker processes this incoming state transition and instantly routes the updated context entities down to the background synchronization nodes of the sensor-deprived asset (AGII). Upon receiving this telemetry, AGII's local navigation stack executes a dynamic costmap injection. As visualized in the RViz interface (Fig.~\ref{fig:rviz_local}), the physical footprints and active safety envelopes of both the peer AGI node and the human grower are dynamically mapped onto AGII's local costmap in real time, successfully overcoming the complete absence of physical, egocentric sensory contact. The evaluation metrics for this simulation are presented in Table \ref{tab:yolo_metrics_sim}.

\begin{figure}[!ht]
    \centering
    \subfloat[Farmer and AGII identification from AGI \label{fig:AGIY}]{
        \includegraphics[width=0.5\textwidth]{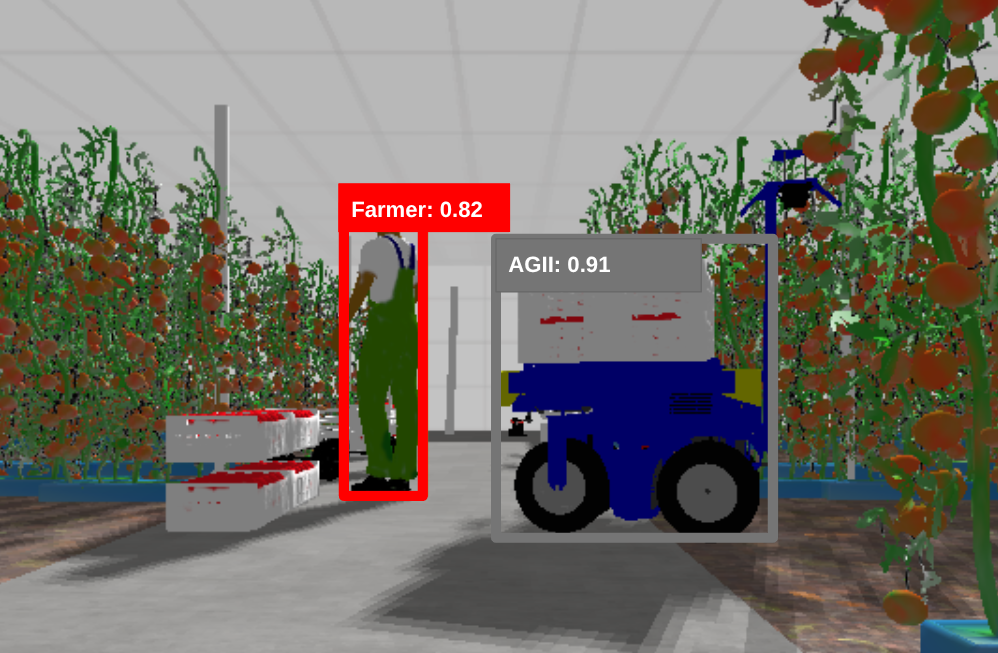}
    }
    \subfloat[Farmer identification from AGI\label{fig:AGIY2}]{
        \includegraphics[width=0.44\textwidth]{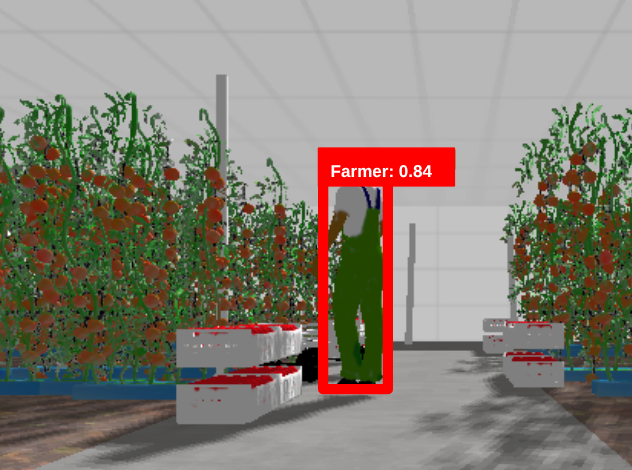}
    }
    \caption{Identification YOLO from AGI}
    \label{fig:YOLOI}
\end{figure}

\begin{figure}[!ht]
    \centering
    \includegraphics[width=0.6\textwidth]{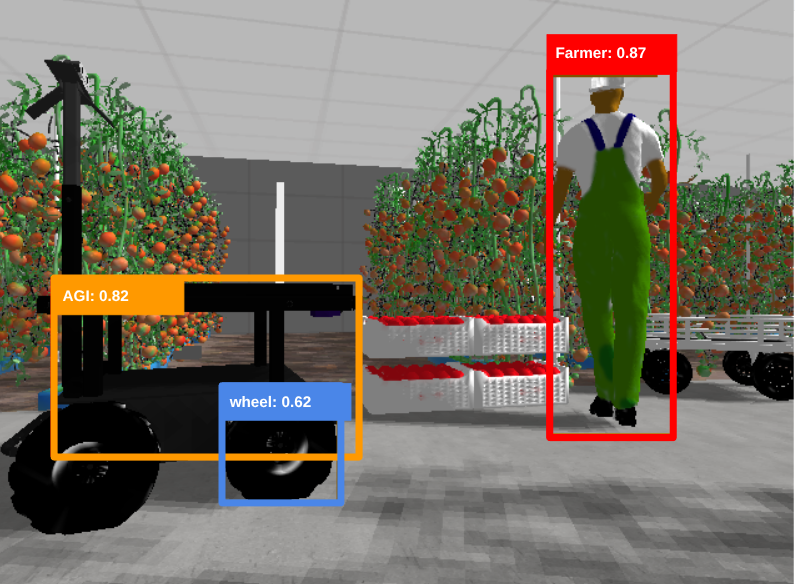}
    \caption{Farmer and AGI identification from AGII}
    \label{fig:yolo_metrics_sim}
\end{figure}

\begin{table}[!ht]
\centering
\caption{Performance metrics of the YOLOX model for object detection in the greenhouse environment (simulation).}
\label{tab:yolo_metrics_sim}
\resizebox{\columnwidth}{!}{
\begin{tabular}{|l|c|c|c|c|c|c|}
\hline
\textbf{Class} & 
\makecell[c]{\textbf{True} \\ \textbf{positive (TP)}} & 
\makecell[c]{\textbf{False} \\ \textbf{positive (FP)}} & 
\makecell[c]{\textbf{False} \\ \textbf{negative (FN)}} & 
\makecell[c]{\textbf{Precision} \\ \textbf{(\%)}} & 
\makecell[c]{\textbf{Recall} \\ \textbf{(\%)}} & 
\textbf{AP@0.5} \\ \hline
Farmer (Human) & 42 & 3 & 1 & 93.33 & 97.67 & 0.94 \\ \hline
Robot AGI & 18 & 2 & 2 & 88.89 & 92.52 & 0.92 \\ \hline
Robot AGII & 28 & 4 & 3 & 87.50 & 87.56 & 0.88 \\ \hline
\rowcolor[HTML]{EFEFEF} 
\textbf{Average (mean)} & - & - & - & \textbf{89.90} & \textbf{92.58} & \textbf{0.91} \\ \hline
\textbf{Total (global)} & \textbf{88} & \textbf{7} & \textbf{6} & \textbf{90.05} & \textbf{92.70} & - \\ \hline
\end{tabular}
}
\end{table}

\begin{figure}[!ht]
    \centering
    \subfloat[MVSim situation \label{fig:mvsim_result2}]{
        \includegraphics[width=0.58\linewidth]{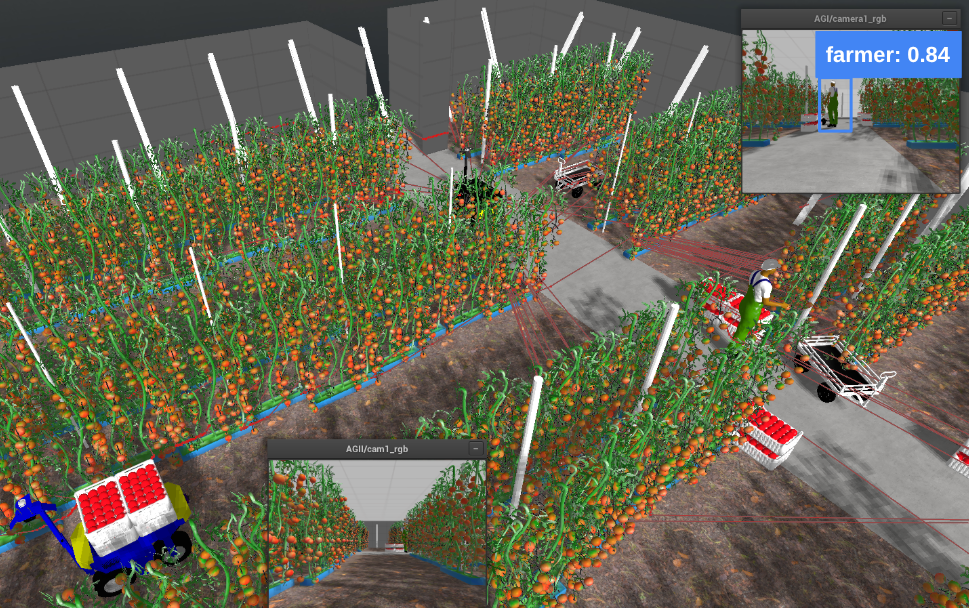}
    }
    \subfloat[Update AGII simulated local map - RViz\label{fig:rviz_local}]{
        \includegraphics[width=0.38\linewidth]{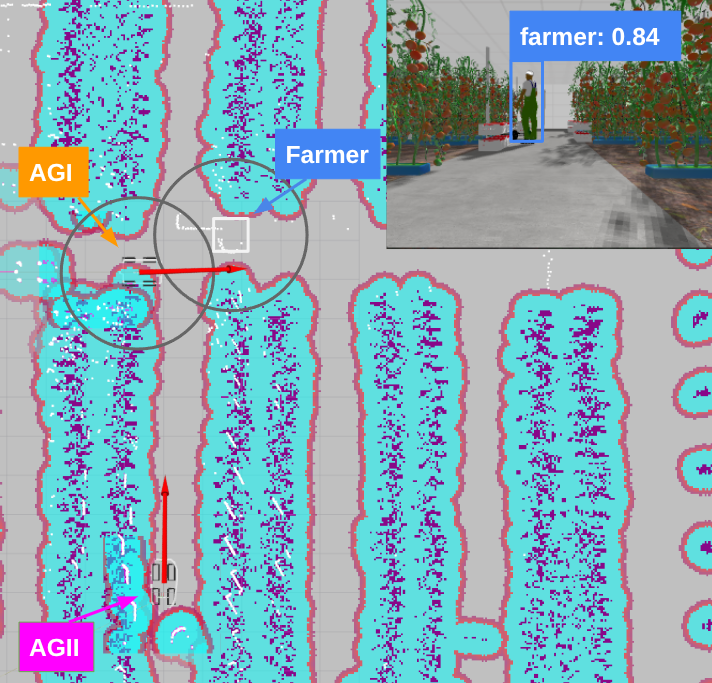}
        }
    \caption{Simulation update AGII local map}
    \label{fig:YOLO_simu2}
\end{figure}

As a direct consequence of this cross-agent synchronization, when the AGII robot advances toward the newly projected warning zone, its local safety node registers the intersection with the human's 2-meter bounding safety envelope. The system triggers an immediate kinematic override, successfully throttling its linear velocity down by 50\% to an operational safety ceiling. This predictive deceleration validates the system's capacity to maintain comprehensive operational safety and human-centric transparency well beyond the immediate, egocentric line of sight of individual robotic assets.

    \item \textit{iVeg Dashboard Multi-Robot Telemetry Contextualization}

    Once the vision nodes were validated, the communication pipeline was integrated with the iVeg application layer (Figure \ref{fig:iVeg}). Rather than acting as a standard engineering telemetry logger, the iVeg interface functions as an Explainable, explicitly decoding the fleet’s internal world-model for the human supervisor through five synchronized visualization widgets:
    
    \begin{itemize}
        \item \textit{Dynamic Semantic 2D Map:} Projects a continuous top-down cartographic representation of the greenhouse lanes. It displays the synchronized poses of the fleet alongside real-time coordinate projections of the AGI (rendered in red) and AGII (rendered in blue) LiDAR point clouds. This global spatial grounding removes the opaqueness of decentralized multi-agent locations.
        
        \item \textit{Kinematic State Feedback:} Extracts and renders the precise global Cartesian spatial coordinates ($X, Y$) and orientation (\textit{Yaw}) variables directly from the OCB's $\mathtt{base\_pose\_ground\_truth}$ digital twin entity.
        
        \item \textit{Proactive Speed Gauges:} Individual visual speedometers map the real-time linear velocities of the fleet in m/s. This widget explicitly displays the immediate physical consequences of the safety overrides when a robot approaches a dynamic boundary.
        
        \item \textit{Action and Inference Log:} A dedicated, chronological telemetry log that couples each YOLOx detection event with its exact timestamp, source asset, target class, and confidence probability. Crucially, it prints the explicit rationale behind any velocity modulation (e.g., displaying $\mathtt{Speed \rightarrow 50\%}$ due to human proximity versus $\mathtt{Speed \rightarrow 100\%}$ upon clearance), granting the user complete, deterministic logical traceability.
        
        \item \textit{Deterministic Emergency Controls:} Standardized interface overrides that allow the supervisor to publish high-priority mechanical $\mathtt{STOP}$ commands via OMA NGSI down to the \texttt{/\{apikey\}/\{device\}/cmd} MQTT channel, providing an immutable human-in-the-loop safety layer.
    \end{itemize}

    To rigorously evaluate the temporal determinism of the communication channel, the experimental data were used to evaluate both central tendency and tail-end dispersion metrics. The mean values quantify the expected steady-state performance, whereas the 95th percentile (P95) metrics isolate transient worst-case latency spikes caused by CPU scheduling contentions or network jitter. The resulting empirical temporal data of the efficacy of the communication framework Table~\ref{tab:mqtt_performance}, where it presents the end-to-end latency decomposition across the full ROS~2--MQTT--FIWARE pipeline, measured over two experimental sessions totalling over 100{,}000 messages. The analysis distinguishes between two latency components: the MQTT transport latency (ROS~2 to the MQTT broker) and the FIWARE processing latency (from the MQTT broker to the Orion Context Broker), providing a detailed view of where delays are introduced in the system. 

    \begin{figure}[!ht]
        \centering
        \includegraphics[width=\textwidth]{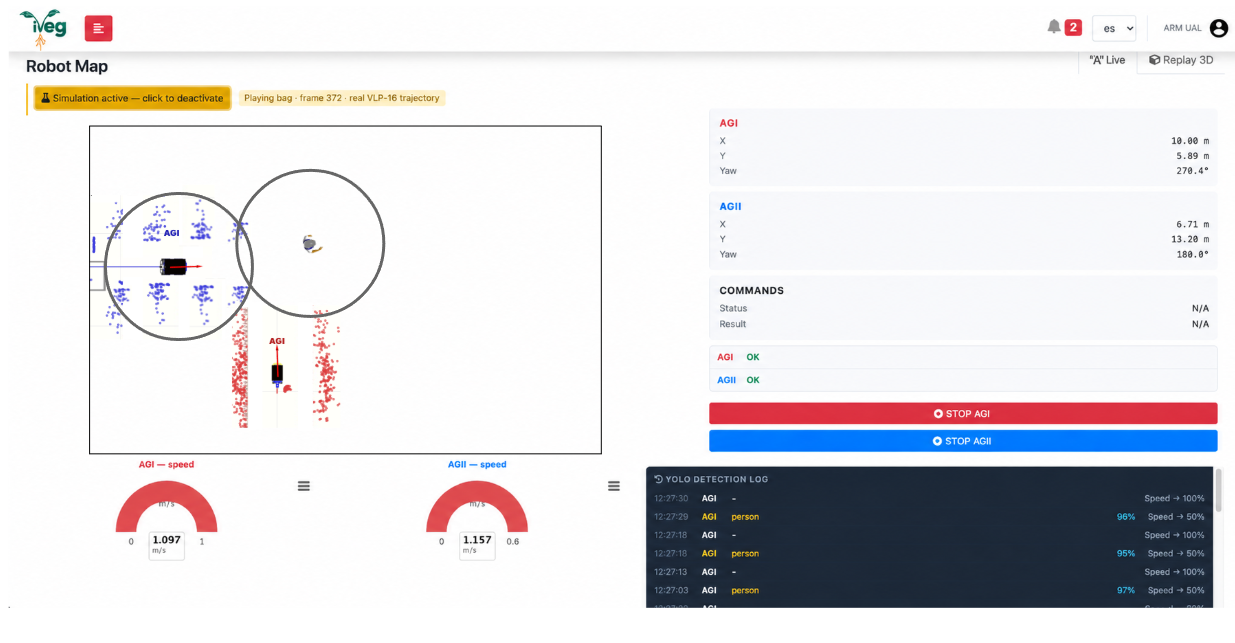}
        \caption{iVeg dashboard for multi-robot monitoring during simulation. The interface shows the 2D greenhouse map with AGI (red) and AGII (blue) LiDAR point clouds and robot trajectories, real-time pose ($X$, $Y$, \textit{Yaw}), individual speed gauges, manual STOP command buttons, and the YOLO Detection Log with confidence scores and speed change events.}
        \label{fig:iVeg}
    \end{figure}
    
    \begin{table}
    \centering
    \caption{End-to-end latency decomposition for the ROS~2--MQTT--FIWARE pipeline in simulation. MQTT transport latency ($\Delta t_1$): ROS~2~$\rightarrow$~MQTT broker (co-located LAN deployment). FIWARE processing latency ($\Delta t_2$): MQTT broker~$\rightarrow$~IoT Agent~$\rightarrow$~Orion Context Broker~$\rightarrow$~MongoDB. Total = $\Delta t_1$ + $\Delta t_2$.}
    \label{tab:mqtt_performance}
    \resizebox{\columnwidth}{!}{%
    \begin{tabular}{|l|c|c|c|c|c|c|}
    \hline
    \textbf{Data Type} & \textbf{$\Delta t_1$ mean} & \textbf{$\Delta t_1$ P95} & \textbf{$\Delta t_2$ mean} & \textbf{$\Delta t_2$ P95} & \textbf{Total mean} & \textbf{Total P95} \\ \hline
    Odometry     & 3.8 ms &  11.2 ms &  98.4 ms & 182.3 ms & 102.2 ms & 193.5 ms \\ \hline
    YOLOx & 4.1 ms &  12.5 ms & 101.2 ms & 188.7 ms & 105.3 ms & 201.2 ms \\ \hline
    LiDAR 2D     & 4.3 ms &  13.1 ms &  96.1 ms & 155.4 ms & 100.4 ms & 168.5 ms \\ \hline
    \end{tabular}}
    \end{table}

    The communication pipeline achieved a 100\% message delivery rate with zero packet dropouts across all simulation cycles. Under Local Area Network (LAN) conditions, the transport overhead ($\Delta t_1$) remained negligible, maintaining a mean of approximately 4~ms. This demonstrates that the serialization logic executing on the ROS~2 edge node introduces no critical computational bottlenecks. The dominant component is the cloud-based processing chain ($\Delta t_2$ mean $\approx$ 99~ms), which encompasses OMA NGSI schema parsing, context entity creation within the Orion Context Broker (OCB), and database commit execution.
    
    From a communication architecture perspective, the global end-to-end P95 latency consistently remains below 202~ms across all semantic data types, including the dense spatial arrays of the YOLOx bounding boxes. This sub-quarter-second latency guarantees real-time visual parity between the physical kinematics of the mobile robots and the graphical representations rendered on the iVeg interface.

\end{itemize}

\subsection{In-Field Validation in the Experimental Greenhouse}

To validate the operational robustness and adaptability of the transparent cloud-robotics framework under unconstrained real-world conditions, both robotic assets were deployed within the physical AgroConnect greenhouse facility. The experimental campaign was designed around a rigorous validation matrix comprising four distinct operational scenarios executed across multiple field runs: i) farmer grower tracking from variable, unconstrained viewpoints; ii) multi-angle inter-agent identification of the robotic peers; iii) simultaneous co-presence of a human operator and a robotic unit within the same camera frame; and iv) multi-agent tracking under severe partial occlusions induced by overlapping tomato crop rows.

The camera intrinsic parameter configurations utilized during the simulation phase were transferred directly to the physical edge nodes. Across the entire experimental campaign, a high operational completion rate of 82\% was achieved, demonstrating the stability of the end-to-end cloud-edge communication pipeline under real agricultural network constraints.

\subsubsection{Real Edge-YOLO validation}

The deep-learning weights optimized during the simulation phase were compiled and deployed onto the onboard processing units of the physical fleets. Within the unstructured physical environment, the perception engine achieved inter-agent and human tracking with a global mean confidence probability of 72\%. Figures \ref{fig:AGI_2_rr} and \ref{fig:AGIIY2_rr} illustrate the real-time inference outputs and bounding box groundings executed by the AGI and AGII assets during the live harvest trials.

The YOLOx model utilized for the detection of farmers and robotic agents was further evaluated using the custom in-field dataset, achieving a global recognition precision of 79.03\% during real-world navigation trials. This performance represents an 11.02\% macro degradation compared to the 90.05\% global precision obtained during idealized simulation baselines. Despite this expected decrease—driven by real-world agricultural challenges such as dynamic solar glare, plastic-diffused lighting, and multi-layered leaf occlusions—these results validate the robustness of the edge-computing perception modules. Even when operating with low-resolution physical sensors, the framework reliably supports safe, collaborative navigation between humans and robots.

From the perspective of the proposed architecture, a cross-sensor comparison reveals that the active RGB-D configuration of the Intel RealSense camera exhibited significantly greater resilience to sudden lighting transients and high-frequency motion blur than the passive Bumblebee2 stereo vision system, which struggled with texture homogeneity against the dense crop backdrop. Nevertheless, the human supervisor can monitor, in real time, both the operational progress of the fleet and the ongoing state transitions occurring within the greenhouse environment.

\subsubsection{Real-world experimental validation updating of the semantic map}

This section examines the multi-robot communication architecture based on a DSS at iVeg in the real greenhouse.

\begin{itemize}
    \item \textit{Experimental upgrading of the local robot semantic map}

    To validate the physical efficacy of the decentralized context-propagation pipeline, an in-field stress test was executed within the physical AgroConnect greenhouse. In this unconstrained environment, physical variables such as shifting foliage canopy density and unexpected structural obstacles introduce real-world processing and network propagation bottlenecks, the specific latency profiles of which are rigorously dissected in the subsequent section. In this experiment, the AGII logistic asset was deployed in the central corridor, deliberately oriented in the opposite direction of the active cultivation rows where human operators were performing harvesting tasks. Simultaneously, the AGI scouting node was positioned directly facing a grower to perform localized edge inference. AGI's perception engine resolved the spatial coordinates of the operator, packaging this semantic metadata alongside its own egocentric odometry profile for immediate cloud publication via the MQTT bridge.

    \begin{figure}[!ht]
        \centering
        \subfloat[Farmer and AGII identification from AGI \label{fig:AGIY_r}]{
             \includegraphics[width=0.35\textwidth]{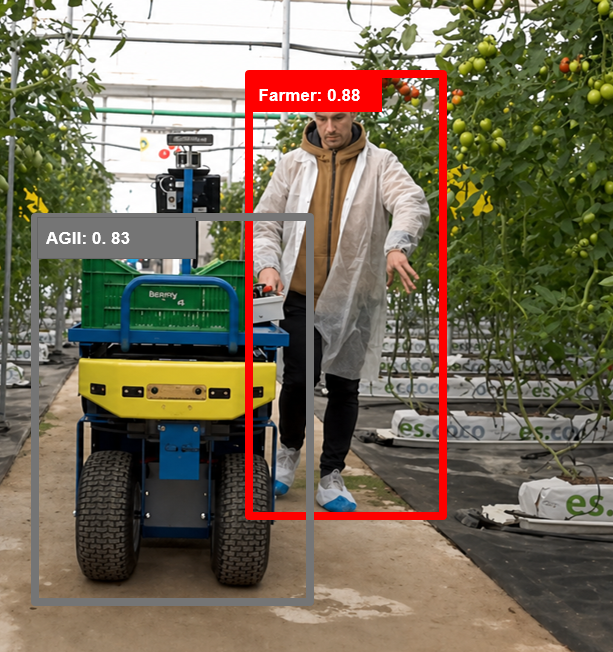}
        }
        \subfloat[Farmer identification from AGI\label{fig:AGIY2_r}]{
            \includegraphics[width=0.55\textwidth]{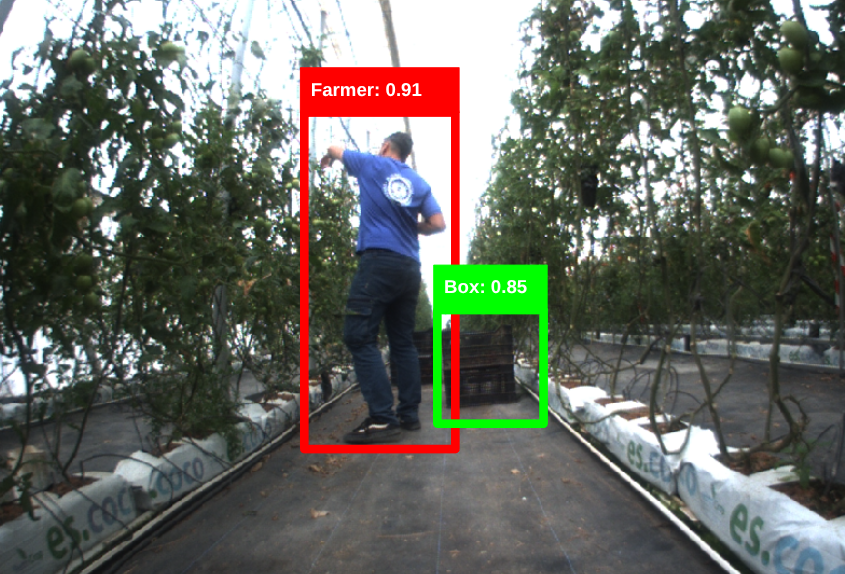}
        }
        \caption{YOLO farmer and AGII recognition from real AGI}
        \label{fig:AGI_2_rr}
    \end{figure}
    
    \begin{figure}[!ht]
        \centering
        \includegraphics[width=0.5\textwidth]{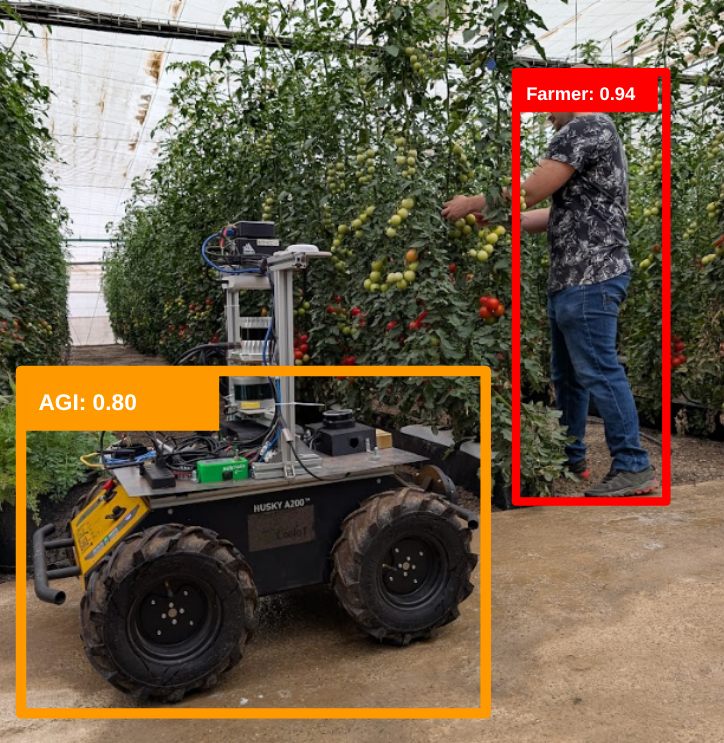}
        \caption{Farmer and AGI identification from AGII}
        \label{fig:AGIIY2_rr}
    \end{figure}

    The results of this live synchronization are detailed in Figure \ref{fig:YOLO_simu} and Table \ref{tab:yolo_metrics_real2}. Figure \ref{fig:mvsim_result} captures the absolute ground-truth physical disposition of the assets alongside the onboard YOLOx detection overlay from the AGI node. Concurrently, Figure \ref{fig:rviz_local2} displays the live state of AGII's local costmap monitored via RViz.

    \begin{table}[!ht]
    \centering
    \caption{Performance metrics of the YOLOX model for object detection in the greenhouse environment (real world).}
    \label{tab:yolo_metrics_real2}
    \resizebox{\columnwidth}{!}{
    \begin{tabular}{|l|c|c|c|c|c|c|}
    \hline
    \textbf{Class} & 
    \makecell[c]{\textbf{True} \\ \textbf{positive (TP)}} & 
    \makecell[c]{\textbf{False} \\ \textbf{positive (FP)}} & 
    \makecell[c]{\textbf{False} \\ \textbf{negative (FN)}} & 
    \makecell[c]{\textbf{Precision} \\ \textbf{(\%)}} & 
    \makecell[c]{\textbf{Recall} \\ \textbf{(\%)}} & 
    \textbf{AP@0.5} \\ \hline
    Farmer (Human) & 36 & 6 & 3 & 85.71 & 92.31 & 0.94 \\ \hline
    Robot AGI & 34 & 8 & 5 & 80.95 & 87.18 & 0.88 \\ \hline
    Robot AGII & 28 & 12 & 8 & 70.00 & 77.78 & 0.88 \\ \hline
    \rowcolor[HTML]{EFEFEF} 
    \textbf{Average (mean)} & - & - & - & \textbf{78.89} & \textbf{85.76} & \textbf{0.90} \\ \hline
    \textbf{Total (global)} & \textbf{98} & \textbf{26} & \textbf{16} & \textbf{79.03} & \textbf{85.96} & - \\ \hline
    \end{tabular}
    }
    \end{table}

    \begin{figure}[!ht]
        \centering
        \subfloat[Real greenhouse situation \label{fig:mvsim_result}]{
        \includegraphics[width=0.35\linewidth]{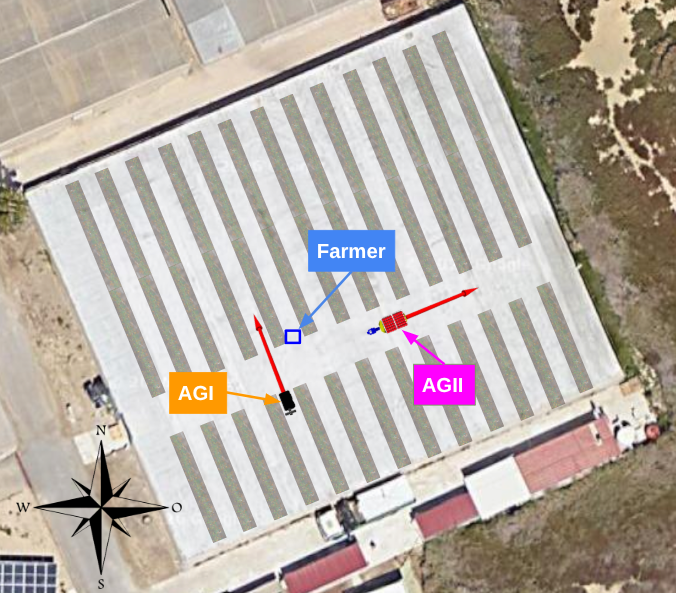}
        }
        \subfloat[Real Update AGII local map - RViz\label{fig:rviz_local2}]{
            \includegraphics[width=0.5\linewidth]{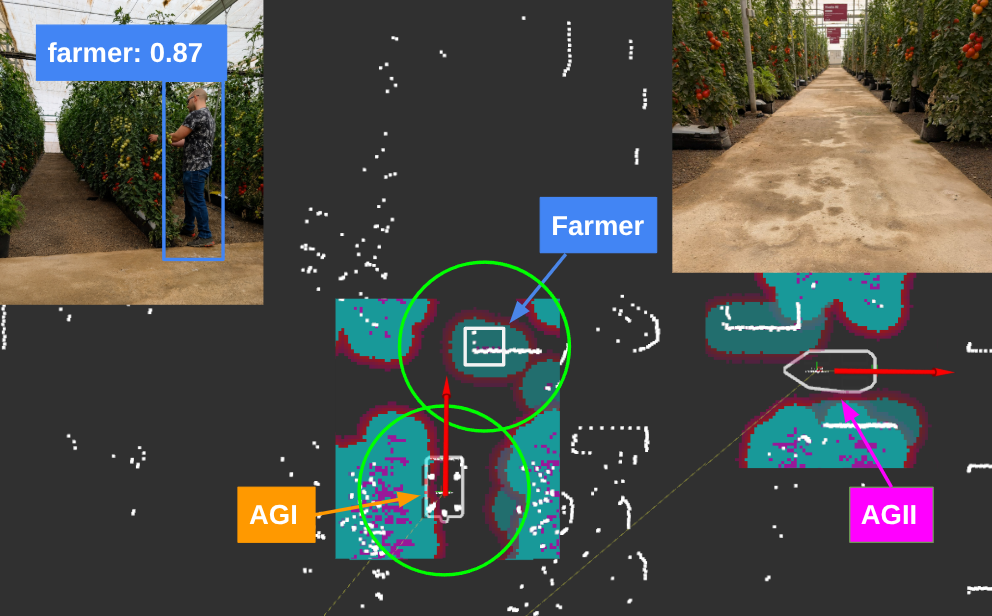}
            }
        \caption{Real update AGII local map}
        \label{fig:YOLO_simu}
    \end{figure}

    As demonstrated by the empirical results, despite AGII's physical optical sensors pointing in the completely opposite direction of the active workspace, the underlying FIWARE-MQTT cloud orchestration layer successfully bypassed its localized line-of-sight limitations. The blind asset seamlessly resolved and rendered the real-time spatial footprints of both its peer unit (AGI) and the human operator within its local navigation costmap. This decentralized awareness eliminates the hazardous opacity of isolated robotic agents, ensuring that even sensor-deprived assets maintain complete, auditable situational awareness of human workers across the shared agricultural environment.
    
\end{itemize}

\begin{itemize}
    \item \textit{iVeg Dashboard multi-robot communication in real greenhouse}

    To guarantee that the iVeg dashboard functions as a dependable, real-time reflection of the fleet's cognitive state, the underlying ROS 2--MQTT--FIWARE communication infrastructure was subjected to a rigorous latency profiling analysis under live greenhouse operating conditions. Maintaining a bounded, low-latency telemetry channel stands as a fundamental requirement. The empirical temporal data logged during active in-field navigation are systematically decomposed in Table \ref{tab:mqtt_performance_real}. This breakdown isolates the transport latency ($\Delta t_1$: edge serialization to the MQTT broker) from the cloud orchestration latency ($\Delta t_2$: broker parsing, OMA NGSI translation via the IoT Agent, Orion Context Broker synchronization, and historical MongoDB persistence).

    \begin{table}
    \centering
    \caption{End-to-end latency decomposition for the ROS~2--MQTT--FIWARE pipeline. MQTT transport latency ($\Delta t_1$): ROS~2~$\rightarrow$~MQTT broker, measured from real robot captures ($n>10{,}000$ per type), clock-offset corrected. FIWARE processing latency ($\Delta t_2$): MQTT broker~$\rightarrow$~IoT Agent~$\rightarrow$~Orion Context Broker~$\rightarrow$~MongoDB, measured via \texttt{TimeInstant} polling ($n=598$, network one-way delay subtracted). Total mean = $\Delta t_1$ mean + $\Delta t_2$ mean.}
    \label{tab:mqtt_performance_real}
    \resizebox{\columnwidth}{!}{%
    \begin{tabular}{|l|c|c|c|c|c|c|}
    \hline
    \textbf{Data Type} & \textbf{$\Delta t_1$ mean} & \textbf{$\Delta t_1$ P95} & \textbf{$\Delta t_2$ mean} & \textbf{$\Delta t_2$ P95} & \textbf{Total mean} & \textbf{Total P95} \\ \hline
    Odometry & 16.9 ms & 103.9 ms & 121.3 ms & 212.6 ms & 138.2 ms & 316.5 ms \\ \hline
    YOLOx    & 16.2 ms & 101.0 ms & 119.2 ms & 214.9 ms & 135.4 ms & 315.8 ms \\ \hline
    LiDAR 2D & 17.4 ms &  97.2 ms & 117.0 ms & 168.6 ms & 134.4 ms & 265.8 ms \\ \hline
    \end{tabular}}
    \end{table}
    
    \begin{figure}[!ht]
        \centering
        \includegraphics[width=\textwidth]{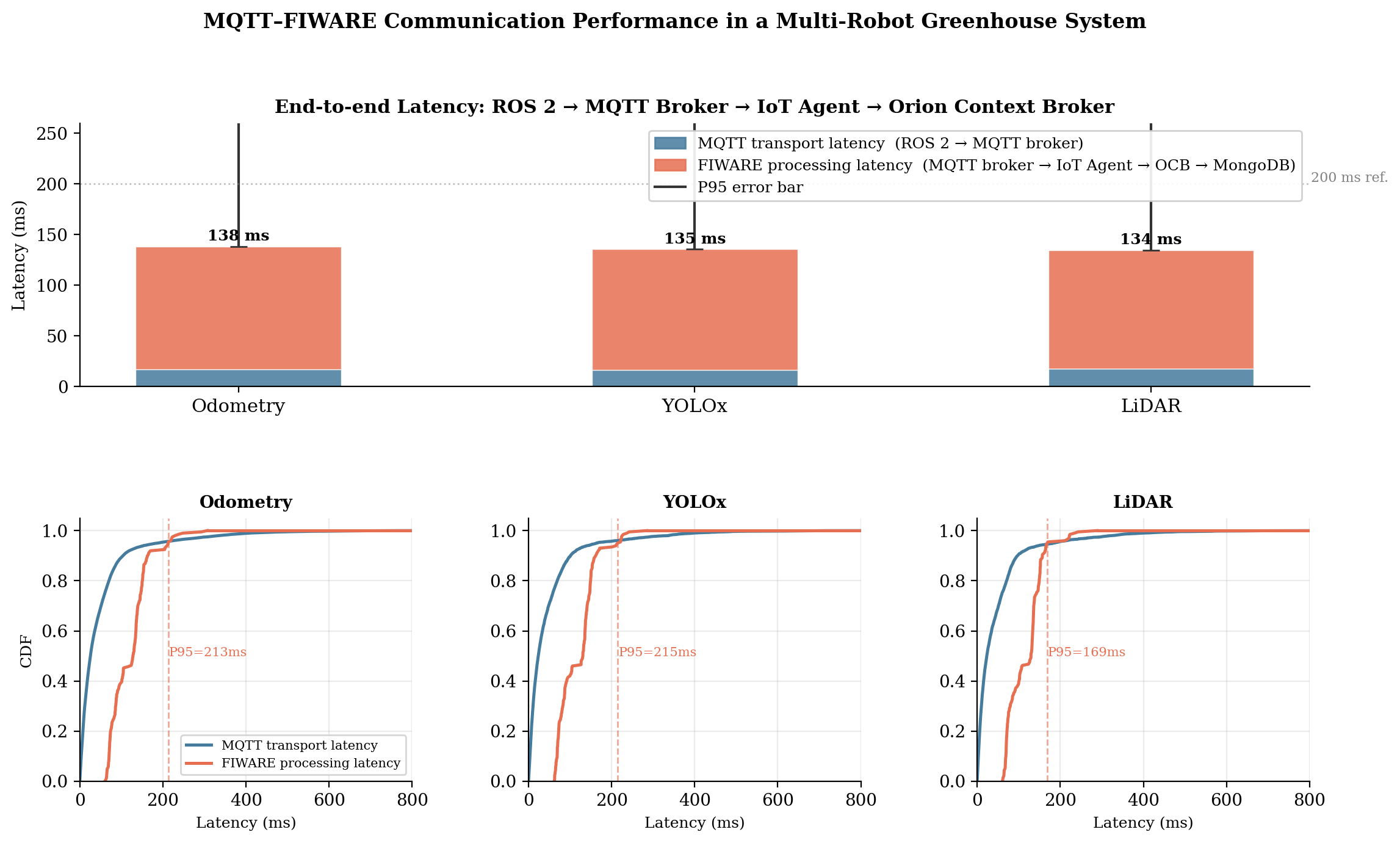}
        \caption{End-to-end latency decomposition across the ROS~2--MQTT--FIWARE pipeline in real environment. Top: mean latency per data type, stacked by MQTT transport (blue) and FIWARE processing (orange), with P95 error bars. Bottom: cumulative distribution functions (CDF) of both latency components per data type.}
        \label{fig:latency_pipeline}
    \end{figure}
    
    As illustrated by the empirical distributions in Figure \ref{fig:latency_pipeline}, the field-measured latency profiles exhibit high consistency with the initial simulation baselines, confirming the structural scalability of the architecture. The edge transport latency ($\Delta t_1$) maintains a remarkably low mean baseline of approximately 17 ms across all semantic payload profiles. This confirms that the lightweight serialization bridges running onboard the mobile assets introduce negligible network overhead, even when operating over agricultural wireless access points.
    
    The observed P95 transport spikes ($\approx 100$ ms) correspond to transient CPU scheduling contentions on the onboard edge computers during moments of peak sensor fusion load, specifically when high-frequency LiDAR registration routines overlap with dense YOLOx coordinate conversions. Rather than hiding these performance fluctuations, the framework's ability to explicitly capture and report these edge constraints demonstrates full operational transparency. The dominant component of the communication channel remains the cloud processing and serialization chain ($\Delta t_2$), yielding a stable mean of approximately 120 ms. This interval accounts for device registration resolution within the IoT Agent, asynchronous NGSI-v2 attribute parsing, context update notifications inside the OCB, and concurrent transactional persistence writes to the MongoDB cluster. Notably, structured numerical matrices (LiDAR arrays) achieved a more tightly bounded processing profile ($\text{P95} = 168.6$ ms) than highly variable metadata streams like Odometry and YOLOx ($\text{P95} \approx 214$ ms). This indicates that standardized, fixed-length payloads are ingested with higher structural efficiency by the cloud translation nodes.
    
    Crucially, the global end-to-end P95 latency consistently remains below 320 ms for every monitored data type under real field operations. In the context of cooperative greenhouse automation, this sub-third-of-a-second threshold easily satisfies the requirements for real-time human-supervised fleet orchestration. This bounded delay guarantees that the real-time event logs, proactive speed gauges, and dynamic map updates rendered on the iVeg user interface remain perfectly synchronized with the physical behaviors of the field units. By eliminating noticeable visual lags or desynchronizations, the system ensures that the human operator is always aware of the exact operational reality of the workspace, establishing a verifiable foundation for human-machine collaboration.
    
\end{itemize}

The following results were drawn within the scope of the real test studies.

\begin{enumerate}
    \item The experimental field trials demonstrate the absolute operational viability of the proposed cloud-robotics and collaborative perception architecture across both simulated and real greenhouse environments. The framework successfully maintained full system transparency and seamless context propagation, enabling safe human-robot cohabitation and achieving an 82\% operational trial completion rate across four highly challenging, unconstrained in-field deployment scenarios.
    
    \item In simulation, the custom-trained YOLOx model established an optimized baseline, achieving a global precision of 90.05\%, a global recall of 92.70\%, and a mean AP@0.5 of 0.91. Transitioning to the unstructured reality of the physical AgroConnect greenhouse introduced severe agricultural noise (diffused plastic glare, dynamic crop shadows, and high-frequency structural vibrations), resulting in a macro-performance degradation where global precision converged to 79.03\%, global recall to 85.96\%, and mean AP@0.5 to 0.90. Crucially, from a human-robot safety standpoint, the critical ``Farmer'' class preserved an exceptionally high recall of 92.31\% in the field. This high sensitivity ensures that human operators are reliably detected by the edge vision nodes, preventing hazardous, sudden ``phantom'' movements and establishing a dependable protective envelope around the human worker.
    
    \item A more detailed per-class analysis reveals notable differences between sensing modalities, directly influencing the reliability of edge-human tracking. The robot equipped with the active RGB-D camera (AGI with Intel RealSense D435) achieves a recall of 87.18\%, whereas the passive stereo-based system (AGII with Bumblebee2) drops to 77.78\%. This suggests that depth sensing using this camera's RGB-D technology provides more robust performance under greenhouse conditions, particularly in the presence of lighting variability and texture-poor regions, thereby offering superior consistency for monitoring human presence and safeguarding cooperative tasks.
    
    \item Regarding the temporal determinism of the transparency communication channel, real-world field profiling validated the high responsiveness of the cloud-robotics architecture. The end-to-end P95 latency remained strictly bounded below 320 ms across all core context data types, while the edge transport delay ($\Delta t_1$) maintained a tight mean baseline of approximately 17 ms. This sub-third-of-a-second performance guarantees real-time visual synchronicity between the physical behaviour of the greenhouse robots and the centralised DSS iVeg. By eliminating noticeable visual lags or out-of-sync telemetry updates, the communication pipeline provides a dependable, verifiable foundation for effective human-in-the-loop supervisor intervention, thereby cementing mutual trust in human-robot collaboration.
\end{enumerate}

\section{Conclusions and Future Work} \label{sec: conclusion}

This paper has presented a field-deployable, high-performance cloud-robo\\tics and multi-agent perception framework designed specifically to optimize data orchestration and edge-computing integration in modern protected agriculture. By coupling an onboard YOLOx semantic inference loop executing as an Edge-Computing routine on each robot’s PC with a standardized ROS 2–MQTT–FIWARE communication pipeline and the iVeg DSS interface, the proposed architecture establishes a robust, highly scalable topology. The system successfully bridges distributed field units with European open-source cloud standards, enabling a fully integrated, low-latency collaborative workspace where non-line-of-sight telemetry is dynamically synchronized across the fleet.

The rigorous field-based validation conducted within the physical AgroConnect greenhouse facility confirms the practical engineering relevance and field-deployability of this cloud-connected architecture. Although transitioning from idealized simulation environments to real-world conditions introduced expected performance degradation due to plastic-diffused glare and complex canopy occlusions—with global perception precision shifting from 90.05\% to 79.03\%—the framework maintained structural and operational integrity. The core of this resilience lies in the decentralization of the computational load, where the MQTT protocol efficiently streams lightweight, high-level semantic metadata parsed from the edge vision nodes, bypassing the need to transmit raw, high-bandwidth sensory data streams to the cloud.

A key insight for the design of modern agricultural networks stems from the clear performance metrics gathered across the telemetry pipeline. By deploying MQTT as the dedicated interoperability layer between local ROS 2 nodes and the FIWARE ecosystem, the end-to-end pipeline operates with a tightly bounded P95 delay under 320~ms, with an edge transport latency ($\Delta t_1$) averaging just 17~ms. This sub-half-second responsiveness guarantees real-time visual synchronicity between physical field kinematics and the graphical updates on the centralized iVeg DSS hub. Furthermore, the active RGB-D sensing modality (Intel RealSense D435) demonstrated greater operational stability against greenhouse lighting fluctuations than the passive stereo system (Bumblebee2), ensuring clean, repeatable inputs for the edge-computing inference layer.

Future work will focus on advancing the scalability and multi-fleet standardization of this ROS 2–MQTT–FIWARE architecture to support larger heterogeneous robotic deployments. Efforts will be directed toward leveraging domain adaptation techniques to close the remaining sim-to-real gap in the edge vision models, alongside retraining the YOLOx nodes with multi-seasonal datasets to further reduce occlusion-driven misclassifications. Additionally, we aim to explore the implementation of advanced MQTT brokers and edge-clustering techniques to maintain bandwidth efficiency within communication-constrained rural infrastructures. Ultimately, by tightly coupling distributed edge intelligence with an open, FIWARE-compliant decision support platform, this research presents a definitive, production-ready blueprint for high-throughput, sustainable multi-robot automation in greenhouse cultivation.

\section*{Acknowledgments} 
This work has been carried out within the framework of the LIFE-ACCLI\\MATE project (LIFE23-CCAES-LIFE-ACCLIMATE/101157315). The first author, Fernando Cañadas-Aránega, holds an FPI grant (PRE2022-102415) from the Spanish Ministry of Science, Innovation, and Universities.
\bibliographystyle{cas-model2-names}

\bibliography{cas-refs}




\end{document}